\documentclass[10pt,twocolumn,letterpaper]{article}

\usepackage{cvpr}              

\usepackage{colortbl}

\definecolor{cvprblue}{rgb}{0.21,0.49,0.74}
\usepackage[pagebackref,breaklinks,colorlinks,allcolors=cvprblue]{hyperref}
\usepackage{multirow}
\usepackage{tabularx}
\usepackage{utfsym}
\usepackage{ragged2e}
\usepackage{comment}
\usepackage{pifont,xcolor}
\usepackage{stfloats} 
\usepackage[normalem]{ulem} 
\usepackage{multirow}
\usepackage{graphicx}
\usepackage{pgfplots}

\newcommand{\bgain}[1]{\begingroup\footnotesize\textcolor{red}{\textbf{#1}}\endgroup} 

\def\paperID{***}
\def\confName{CVPR}
\def\confYear{2026}

\title{SPARROW \includegraphics[height=2ex]{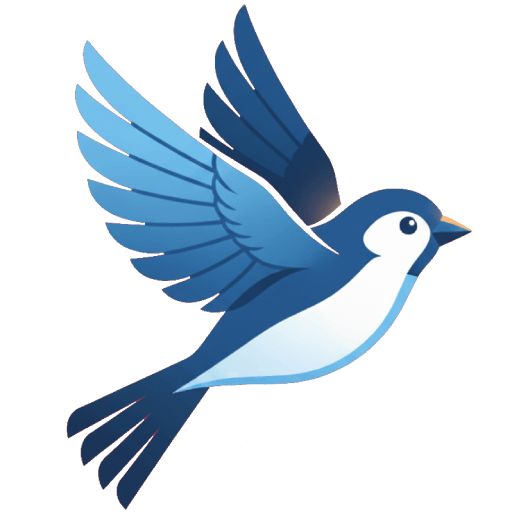}: Learning Spatial Precision and Temporal Referential \\ Consistency in Pixel-Grounded Video MLLMs}

\usepackage{risys-title}

\RisysTitle{SPARROW \includegraphics[height=2ex]{figures/sparrow.png}: Learning Spatial Precision and Temporal Referential Consistency in Pixel-Grounded Video MLLMs}
\RisysAuthor[*]{Mohamad Alansari}
\RisysAuthor[*]{\quad Naufal Suryanto}
\RisysAuthor[]{\quad Divya Velayudhan}
\RisysAuthor[]{\quad Sajid Javed}
\RisysAuthor[]{\\Naoufel Werghi}
\RisysAuthor[]{\quad Muzammal Naseer}
\RisysAffil[]{Khalifa University, Abu Dhabi, UAE}
\RisysContribution[*]{Equal contributors}
\RisysProjectpage{https://risys-lab.github.io/SPARROW}
\RisysRepository{https://github.com/RISys-Lab/SPARROW}

\RisysAbstract{
\vspace{-3mm}
Multimodal large language models (MLLMs) have advanced from image-level reasoning to pixel-level grounding, but extending these capabilities to videos remains challenging as models must achieve spatial precision and temporally consistent reference tracking. 
Existing video MLLMs often rely on a static segmentation token ([SEG]) for frame-wise grounding, which provides semantics but lacks temporal context, causing spatial drift, identity switches, and unstable initialization when objects move or reappear. 
We introduce \textbf{SPARROW}, a pixel-grounded video MLLM that unifies spatial accuracy and temporal stability through two key components: (i) Target-Specific Tracked Features (TSF), which inject temporally aligned referent cues during training, and (ii) a dual-prompt design that decodes box ([BOX]) and segmentation ([SEG]) tokens to fuse geometric priors with semantic grounding. 
SPARROW is supported by a curated referential video dataset of 30{,}646 videos and 45{,}231 Q\&A pairs and operates end-to-end without external detectors via a class-agnostic SAM2-based proposer. 
Integrated into three recent open-source video MLLMs (UniPixel, GLUS, and VideoGLaMM), SPARROW delivers consistent gains across six benchmarks, improving up to +8.9 $\mathcal{J\&F}$ on RVOS, +5 mIoU on visual grounding, and +5.4 CLAIR on GCG. 
These results demonstrate that SPARROW substantially improves referential stability, spatial precision, and temporal coherence in pixel-grounded video understanding. Code, datasets, and models are released.
\vspace{-3mm}
}

\makeatletter
\renewcommand{\paragraph}{\@startsection{paragraph}{4}{\z@}%
  {0.5ex plus 0.1ex minus 0.1ex}
  {-1em}
  {\normalfont\normalsize\bfseries}}
\makeatother

\begin{document}
\RisysMakeTitle
\begin{figure*}[!t]
    \centering

    \begin{subfigure}[b]{0.49\textwidth}
        \includegraphics[width=\textwidth]{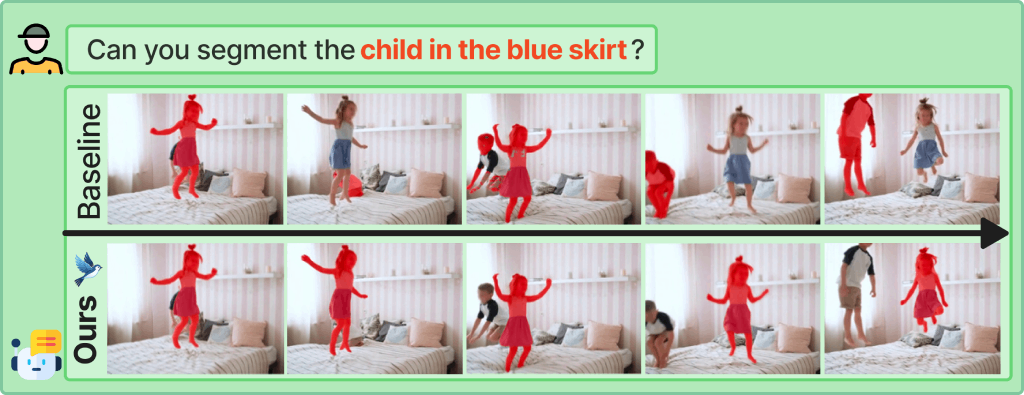}
        \caption{Issue: Inconsistent Referencing}
        \label{fig:intro_a}
    \end{subfigure}
    \hfill
    \begin{subfigure}[b]{0.49\textwidth}
        \includegraphics[width=\textwidth]{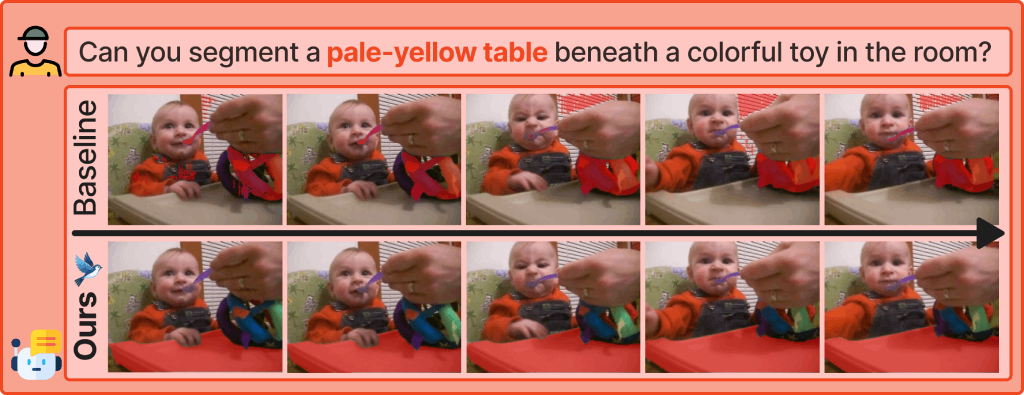}
        \caption{Issue: Noisy Initialization}
        \label{fig:intro_b}
    \end{subfigure}
    
    \begin{subfigure}[b]{0.49\textwidth}
        \includegraphics[width=\textwidth]{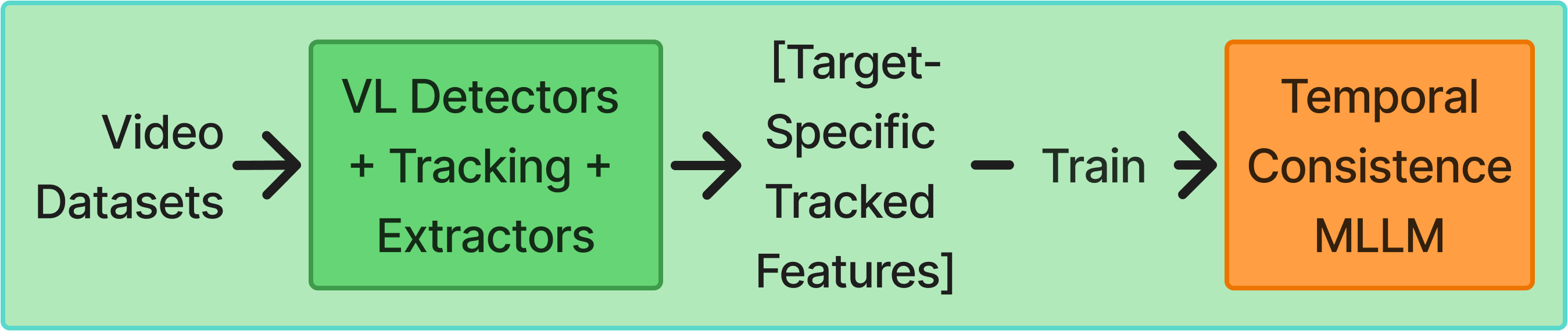}
        \caption{Our solution: Target-Specific Tracked Feature (TSF)}
        \label{fig:intro_c}
    \end{subfigure}
    \hfill
    \begin{subfigure}[b]{0.49\textwidth}
        \includegraphics[width=\textwidth]{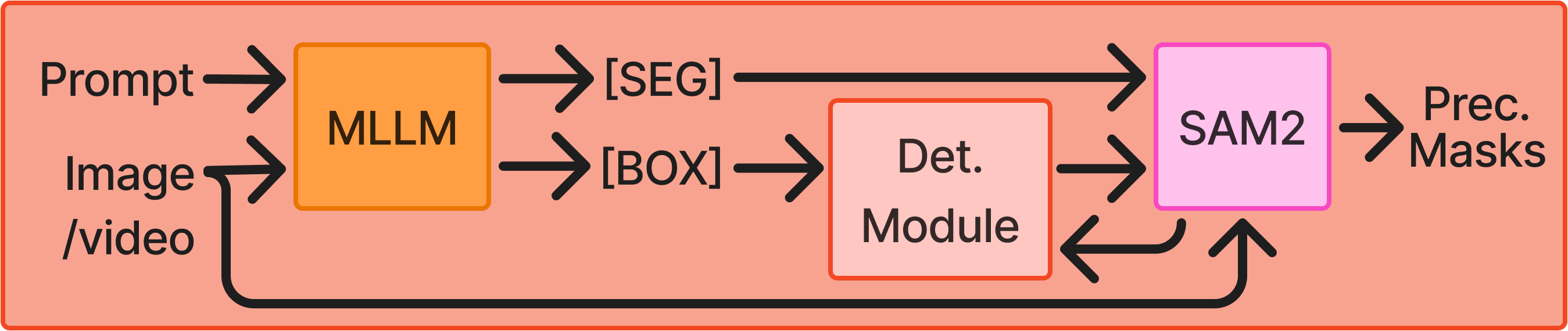}
        \caption{Our solution: Dual Prompt Initialization}
        \label{fig:intro_d}
    \end{subfigure}
    \caption{Comparison of temporal consistency and initialization quality in video object segmentation. 
    (a) The baseline method~\cite{videoglamm} suffers from temporal drift, leading to inconsistent segmentation of the same object across frames. 
    (b) Noisy or unstable initialization propagates segmentation errors through subsequent frames. 
    (c) Our proposed \emph{Target-Specific Tracked Feature} mitigates drift by maintaining consistent object grounding over time. 
    (d) The \emph{Dual-Prompt Initialization} strategy improves segmentation precision and stability during early frames.
    }
    \label{fig:intro}
\end{figure*}

\section{Introduction}
\label{sec:intro}

Recent advances in Multimodal Large Language Models (MLLMs) have enabled remarkable progress in visual reasoning across text and vision, ranging from visual question answering~\cite{llava,lisa} to pixel-level grounding in images~\cite{kosmos2,groundingdino,gpt4roi,position_enhanced}. 
These models bridge perception and language understanding by aligning visual entities with linguistic references. 
However, extending image-grounded MLLMs to the video domain remains challenging because videos introduce motion dynamics, occlusion, and the need for temporally coherent grounding.

Despite progress in multimodal reasoning, Video MLLMs  still struggle to maintain object identity and spatial coherence across frames (Fig.~\ref{fig:intro_a}). 
Most existing approaches rely on static textual grounding tokens (e.g., \texttt{[SEG]})~\cite{videoglamm,videomolmo,sa2va,interrvos,unipixel,svac,sama,glus}, which indicate what to look for but provide no information about how an object’s position or appearance evolves over time. 
Because text prompts are static while videos are dynamic, the model must infer motion and appearance changes entirely from visual cues, which often leading to spatial drift, inconsistent referent tracking, and unstable segmentation when the same entity moves or reappears.
%
%
These issues are further emphasized by unreliable first-frame initialization (Fig.~\ref{fig:intro_b}). 
Since the \texttt{[SEG]} token supplies only semantic cues without spatial priors, the initial mask may frequently misaligns with the target, and such errors accumulate as the sequence progresses. 
Once this drift begins, object identity switches and referential incoherence follow, degrading temporal stability and limiting performance in complex video question answering and referring video segmentation.


To address these challenges, we propose \textbf{SPARROW} (\textbf{S}patial \textbf{P}recision and \textbf{R}eferential \textbf{R}easoning in \textbf{O}bject-centric \textbf{\sout{W}}Video grounding), 
a pixel-grounded, temporally consistent Video MLLM that enhances referential stability and first-frame robustness. 
As illustrated in Fig.~\ref{fig:pipeline}, SPARROW jointly learns temporal consistency and spatial precision through two complementary components.

\textbf{(1) Target-Specific Tracked Feature (TSF).}
To enforce temporal referential consistency, we introduce a TSF mechanism that tracks object instances across frames and provides temporally aligned features during training (Fig.~\ref{fig:intro_c}). 
This design allows the model to maintain identity and spatial coherence over time, enabling stable temporal representation learning directly from tracked visual signals. 
To support TSF supervision, we construct a new large-scale dataset, tailored for referential video understanding and fine-grained segmentation. 
Unlike existing video datasets~\cite{vidstg,videoglamm} that emphasize global comprehension or scene-level captioning, our dataset provides object-centric, temporally consistent annotations essential for training target-specific tracking and grounding. 
It comprises 30,646 video sequences and 45,231 Q\&A pairs with high-quality bounding boxes, segmentation masks, and temporally aligned trajectories.

\textbf{(2) Dual-Prompt Initialization.}
We further introduce a Dual-Prompt strategy that integrates \texttt{[BOX]} and \texttt{[SEG]} tokens during both training and inference to stabilize geometric and semantic grounding (Fig.~\ref{fig:intro_d}). 
The \texttt{[BOX]} token provides a coarse spatial prior by conditioning a lightweight regression head on class-agnostic region proposals extracted from SAM2’s multi-scale features \cite{sam2, hiera}, enabling geometry-aware localization without external detectors. 
The \texttt{[SEG]} token then refines these regions through language-conditioned semantics to produce precise masks. 
By jointly decoding the two prompts, the model learns to align spatial priors and semantic cues, improving convergence and reducing early-frame ambiguity.


By combining temporally supervised training with dual-prompt grounding, SPARROW achieves stable, coherent video understanding running end-to-end without relying on external detectors/trackers. 
Its modular design seamlessly adapts to three state-of-the-art video MLLMs \cite{unipixel,videoglamm,glus}, yielding consistent gains across six benchmarks. 

\section{Related Work}
\label{sec:related_work}

\subsection{From Language Models to Grounded MLLMs}
\paragraph{Language and Multimodal Foundations.}
Large Language Models (LLMs) demonstrate strong reasoning and instruction-following abilities \cite{language_few_shot_learner,scaling,lamda,palm,opt,llama,glm,vicuna}. 
Building on this foundation, MLLMs integrate visual and linguistic information through internal cross-attention \cite{flamingo} or adapters that connect frozen vision encoders to language models \cite{blip2,instructblip,llava}. 
With high-capacity vision transformers \cite{vit,swin,clip,int_pre_trained,tiansemantic,hivit,sam}, these models are able to perceive detailed visual information and support instruction-grounded perception  \cite{llava,tianchatterbox,gpt4roi,lisa}. 

\paragraph{Grounded and Referring Understanding in Images.}
Grounded MLLMs extend beyond global QA to region- and pixel-level reasoning \cite{kosmos2,groundingdino,gpt4roi,position_enhanced}. 
Spatial cues are modeled explicitly with position tokens or coordinates \cite{kosmos2,visionllm}, or implicitly through language \cite{shikra,cogvlm,pink}. 
Recent models couple language reasoning with dense segmentation \cite{lisa,glamm}, achieving fine-grained grounding on static content. 
However, these methods operate on images and therefore do not maintain object identity or coherence across time.


\subsection{Video MLLMs for Grounded Segmentation}
\paragraph{Video MLLMs.}
Extending grounded MLLMs from images to videos introduces long‑range temporal dynamics and occlusion \cite{videochat,videochatgpt,videollama,videollava,languagebind,valley}. Early works emphasize holistic comprehension or moment retrieval \cite{groundinggpt}, and later systems compose tracking with grounding \cite{pgvideollava}. Recent pixel‑grounded video MLLMs \cite{videoglamm,videomolmo,sa2va,interrvos,unipixel,svac,sama,glus} advance language‑conditioned segmentation, yet most still rely on a static \texttt{[SEG]} token for frame‑wise grounding. In particular, VideoGLaMM \cite{videoglamm} prompts a SAM‑style decoder via \texttt{[SEG]} without explicit temporal cues, UniPixel \cite{unipixel} unifies mask‑grounded reasoning with an online memory seeded by first‑frame masks, and GLUS \cite{glus} couples global context with dense query frames using a trainable memory. 
Despite strong results, these designs rely on per-frame semantics and propagated masks instead of sequence-level referential cues, making them prone to drift, identity switches, and fragile re-initialization.

\paragraph{Our Positioning.}
We address these limitations by introducing two complementary components that jointly model temporal and spatial grounding cues. First, inspired by Artemis \cite{artemis}, which shows that tracking object‑specific features improves temporal coherence, we introduce target‑specific tracked features (TSF) to maintain referential consistency across frames. Second, building on localized box prompting from Groma \cite{groma}, which enhances fine‑grained visual grounding in images, we design a dual‑prompt mechanism for video understanding that unifies \texttt{[BOX]} and \texttt{[SEG]} within a coarse‑to‑fine framework, integrating spatial priors with semantic cues during both training and inference. 
Our framework is plug-and-play and attaches to video MLLMs through lightweight adapters without requiring backbone modification. To demonstrate its practicality and generality, we integrate SPARROW into three recent open-source Video MLLMs \cite{unipixel,glus,videoglamm}
and observe consistent improvements in temporal stability and spatial precision across grounded segmentation tasks.

\begin{figure*}[!t]
    \centering
    \includegraphics[width=\linewidth]{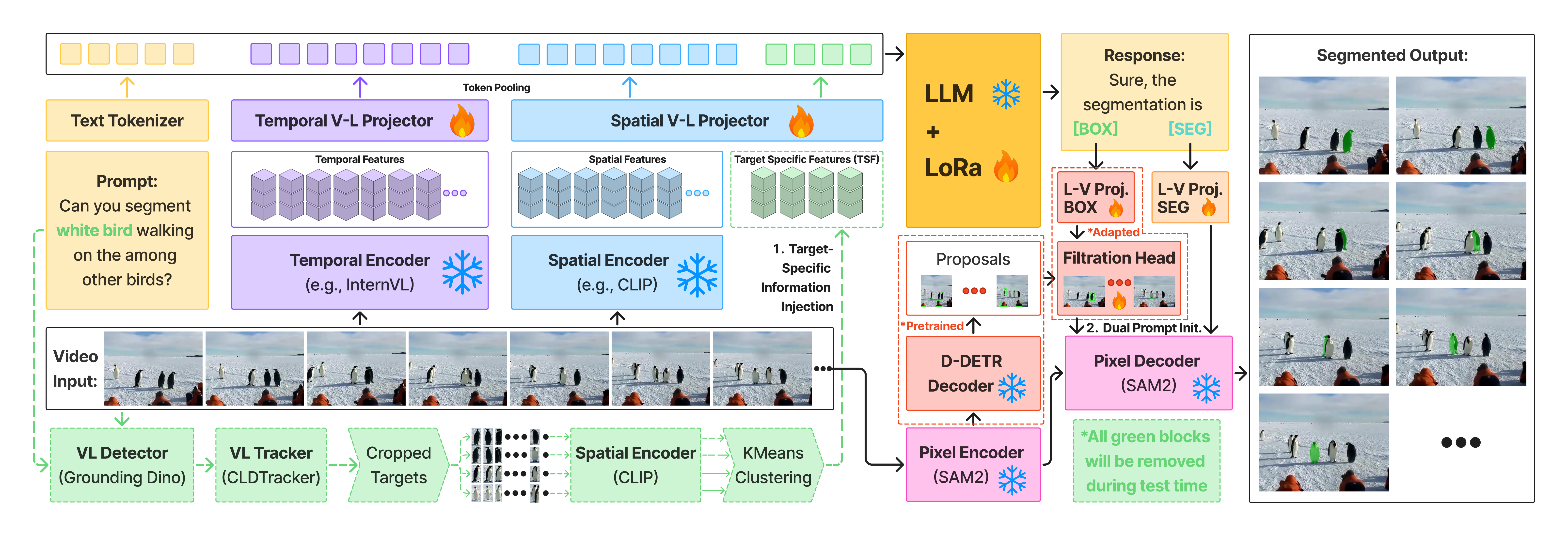}
    \vspace{-2em}
    \caption{\textbf{SPARROW pipeline.} Given a video and text prompt, spatial \cite{clip} and temporal \cite{internvl2} encoders feed V$\!\to$L adapters and a LoRA‑tuned LLM. The LLM emits \texttt{[BOX]} and \texttt{[SEG]}  tokens which are projected (L$\!\to$V) to condition a class‑agnostic proposer and the SAM2 \cite{sam2} pixel decoder. Dashed green modules (GroundingDINO \cite{groundingdino}, CLDTracker \cite{cldtracker}, target cropping, K‑means) are pre-computed offline as pseudo-supervision, used only for \emph{target-specific information injection step}, and are removed at test time by default.}
    \label{fig:pipeline}
\end{figure*}

\section{Methodology} \label{sec:method}

\subsection{Framework Overview} \label{sec:framework}
We address two key limitations of pixel-grounded video MLLMs \cite{videoglamm,videomolmo,sa2va}: 
(i) \emph{temporal referential consistency} (identity switches), and 
(ii) \emph{error propagation from unstable first-frame grounding} (drift). 
SPARROW augments a baseline video MLLM with two complementary modules (Fig.~\ref{fig:pipeline}): 
(1) Target-Specific Features (TSF), which inject temporally aligned referent cues during training, and 
(2) Dual-Prompt Grounding, which decodes \texttt{[BOX]} and \texttt{[SEG]} tokens to couple geometry with semantics. 
At a high level, \texttt{[BOX]} conditions a class-agnostic proposer built on SAM2/Hiera features \cite{sam2,hiera} to produce bounding boxes, while \texttt{[SEG]} provides language-conditioned semantics; together they guide SAM2’s decoder in a coarse-to-fine manner, stabilizing early frames and mitigating drift.

\paragraph{Architecture.}
The framework comprises: (i) a dual-branch visual encoder for spatial \cite{clip} and temporal \cite{internvl2} features, (ii) Visual-to-Language (V$\!\to$L) adapters, (iii) a frozen LLM adapted via LoRA \cite{lora}, (iv) Language-to-Visual (L$\!\to$V) adapters that map grounding states back to pixels, and (v) a SAM2-based pixel segmentor \cite{sam2}. All additional modules are plug-and-play and do not alter the base LLM or visual backbones.

\paragraph{Visual Feature Encoding.}
Given a video $\mathbf{V}\!\in\!\mathbb{R}^{T_{\!v}\times H\times W\times C}$, an image encoder $\mathcal{F}_g$ and a video encoder $\mathcal{F}_h$ extract spatial and temporal features:
\begin{equation}
    f_g = \mathcal{F}_g(\mathbf{V}), \qquad f_h = \mathcal{F}_h(\mathbf{V}).
\end{equation}
These features are projected into the LLM embedding space via the V$\!\to$L adapters:
\begin{equation}
    Z_g = \mathcal{W}_g(f_g), \qquad Z_h = \mathcal{W}_h(f_h).
\end{equation}
Given a tokenized query $Q\!\in\!\mathbb{R}^{N_t\times D}$, the multimodal input is defined as:
\begin{equation}
    \mathcal{I} = [\,Q;\,Z_g;\,Z_h;\,Z_{\text{TSF}}\,],
    \label{eq:multimodal_input}
\end{equation}
where $Z_{\text{TSF}}$ (see Sec.~\ref{sec:tsf}) represents optional TSF tokens introduced during training. 
Grounding-related hidden states are projected back into the visual space through L$\!\to$V adapters $\mathcal{W}_b$ (for \texttt{[BOX]}) and $\mathcal{W}_s$ (for \texttt{[SEG]}), enabling the model to decode spatial and semantic prompts simultaneously. 
Detailed mechanisms for dual-prompt grounding are presented in Sec.~\ref{subsec:dualprompt}.

\subsection{Target-Specific Features} \label{sec:tsf}
\paragraph{Motivation.}
Video MLLMs often encode holistic scene semantics but struggle to maintain object identity across frames. TSF injects temporally aligned, referent-specific cues \emph{during training}, enabling identity persistence without explicit tracking at inference.

\paragraph{Tracking and Selection.}
Given a textual query $Q$, GroundingDINO \cite{groundingdino} detects the referent in one frame, and CLDTracker \cite{cldtracker} propagates it across the sequence, producing candidate boxes 
$\mathcal{B}'=\{\mathbf{B}'_1,\ldots,\mathbf{B}'_{K'}\}$, 
where $K'$ denotes the total number of tracked boxes. 
To reduce redundancy and improve stability, we perform K\mbox{-}means clustering in a joint visual–spatial feature space with $K$ clusters and retain the samples nearest to each centroid, resulting in a compact subset 
$\mathcal{B}=\{\mathbf{B}_1,\ldots,\mathbf{B}_K\}$ with $K<K'$. 
Following the insights of Artemis~\cite{artemis}, this selection ensures that each $\mathbf{B}_j$ represents a diverse appearance of the same referent identity.

\paragraph{Feature Encoding and Integration.}
We encode tracked regions, select representatives via K\mbox{-}means (we set $K=4$), and project the resulting features into the LLM space:
\begin{equation}
    \underbrace{
        F_{\text{TSF\_all}} 
        = \big\{\mathcal{F}_g(\mathbf{B}'_i)\big\}_{i=1}^{K'}
    }_{\text{encode all tracked boxes}}
    \xrightarrow{\;\text{K-means}\;}
    \underbrace{
        F_{\text{TSF}} 
        = \big\{\mathcal{F}_g(\mathbf{B}_j)\big\}_{j=1}^{K}
    }_{\text{selected regions}},
\end{equation}
\begin{equation}
    Z_{\text{TSF}}
    = \big\{
        \mathcal{W}_g(f)
        \mid
        f \in F_{\text{TSF}}
      \big\}.
\end{equation}
$Z_{\text{TSF}}$ forms the TSF tokens appended to the input as in Eq.~\ref{eq:multimodal_input}.
This lightweight integration supplies temporally consistent referent cues while leaving the base architecture unchanged.

\paragraph{Offline Dataset Curation.}
To support TSF supervision, we curate a large-scale dataset tailored for referential video understanding and fine-grained segmentation. 
Existing datasets~\cite{vidstg,videoglamm} primarily focus on holistic scene comprehension rather than object-centric temporal grounding. 
We therefore harmonize multiple publicly available sources, including HC-STVG~\cite{hcstvg}, VID-Sentence~\cite{vidsentence}, A2D Sentences~\cite{a2d}, LaSOT~\cite{lasot}, MeViS~\cite{mevis}, GOT-10k~\cite{got10k}, and Ref-SAV~\cite{sa2va,sam2}, into a unified corpus. 
To enable efficient training, we precompute detections and trajectories, encode features, and prune them using K-means clustering~\cite{artemis}, optionally generating dense masks with SAM2~\cite{sam2}. 
The resulting dataset contains 30{,}646 video sequences and 45{,}231 question–answer pairs, providing temporally consistent trajectories, bounding boxes, and dense segmentation masks. 
This offline pipeline decouples heavy modules from the training loop while preserving temporal supervision benefits. 
Additional statistics and details are in Appendix A. 


\paragraph{Inference.}
TSF tokens are optional at test time. By default, we omit them (no external detector or tracker). When enabled, we execute GroundingDINO $\!\to$ CLDTracker $\!\to$ CLIP $\!\to$ K-means to inject referent cues for maximal temporal stability. We report the trade-off between using and omitting TSF in ablation studies (Sec. ~\ref{Ablation-Study}).

\subsection{Dual-Prompt Grounding} \label{subsec:dualprompt}

\paragraph{Motivation.}
Early segmentation errors can cascade into drift and identity switches. 
We address this with \emph{dual prompts}: \texttt{[BOX]} for spatial priors and \texttt{[SEG]} for semantic grounding. Together, they enable coarse-to-fine reasoning that stabilizes early frames and allows recovery from drift.


\paragraph{Bounding-box Prompt.}
The LLM emits a \texttt{[BOX]} token with embedding $e_{\text{BOX}}\!\in\!\mathbb{R}^d$ (projected by $\mathcal{W}_b$) that conditions a lightweight regression head to predict $(x_1,y_1,x_2,y_2)$. 
We build a class-agnostic proposer on frozen Hiera features from SAM2 \cite{sam2,hiera}. 
Following \cite{vitdet,groma}, multi-scale Hiera maps are aligned into a feature pyramid and fed to a Deformable-DETR decoder \cite{deformable_detr}, replacing the classification branch with a single objectness head. 
This yields $K{=}300$ proposals per frame, filtered by objectness and NMS to retain high-confidence anchors.

\noindent \textit{Language-conditioned refinement.}
For each proposal $b_i$, we pool Hiera features with $\mathrm{ROIAlign}$ and aggregate:
\begin{equation}
    \small
    \{F_i^\ell\}_{\ell=1}^{L}=\mathrm{ROIAlign}\!\left(\{F^\ell\}_{\ell=1}^{L},\, b_i\right),
    \quad
    G_i=\mathrm{Pool}\!\left(\{F_i^\ell\}_{\ell=1}^{L}\right).
\end{equation}
We then fuse $e_{\text{BOX}}$ with per-proposal features by cross-attention:
\begin{equation}
    A_i=\mathrm{softmax}\!\left(\frac{(W_q e_{\text{BOX}})(W_k F_i)^\top}{\sqrt{d}}\right),
    \quad
    Z_i=A_i(W_v F_i).
\end{equation}
Spatial pooling over $Z_i$ yields a compact cross-modal descriptor $Z_i^{\text{pool}}$. We fuse and score:
\begin{equation}
    v_i=[\,Z_i^{\text{pool}};\,G_i;\,e_{\text{BOX}}\,], 
    \quad
    s_i=\sigma\!\big(W_2\,\mathrm{ReLU}(W_1 v_i)\big).
\end{equation}
For the top-$M$ candidates, a text-conditioned regression head refines each box:
\begin{equation}
    \Delta b_i=f_{\text{ref}}\!\big([Z_i^{\text{pool}};\,G_i;\,e_{\text{BOX}}]\big),
    \quad
    \hat b_i=b_i \oplus \Delta b_i.
\end{equation}
Final confidence combines language and visual cues:
\begin{equation}
    s_i^{\text{final}}=\sigma\!\big(\alpha\, s_i+\beta\, \mathrm{logit}(p_i^{\text{det}})\big),
    \quad
    \mathcal{B}^{\star}=\{\,\hat b_i \mid s_i^{\text{final}}>\tau\,\}.
\end{equation}
Each $\hat b_i\!\in\!\mathcal{B}^{\star}$ provides a spatial prior that constrains subsequent segmentation, and independent scoring naturally supports multi-instance queries (e.g., “both players”) without explicit instance supervision.

\paragraph{Segmentation Prompt.}
The LLM also emits a \texttt{[SEG]} embedding $e_{\text{SEG}}\!\in\!\mathbb{R}^d$ (projected by $\mathcal{W}_s$) to SAM2’s prompt encoder. 
For each $\hat b_i\!\in\!\mathcal{B}^{\star}$, the pair $(\hat b_i, e_{\text{SEG}})$ forms a mask query to SAM2, producing a segmentation $\hat{\mathbf{M}}_i$. 
When $|\mathcal{B}^{\star}|{>}1$, SAM2 returns instance masks for each spatial prior, naturally supporting multi-instance outputs. 
Re-emitting \texttt{[BOX]} and \texttt{[SEG]} at arbitrary frames enables drift correction without external detectors or trackers.

\begin{figure*}[t]
  \centering
  \includegraphics[width=\textwidth]{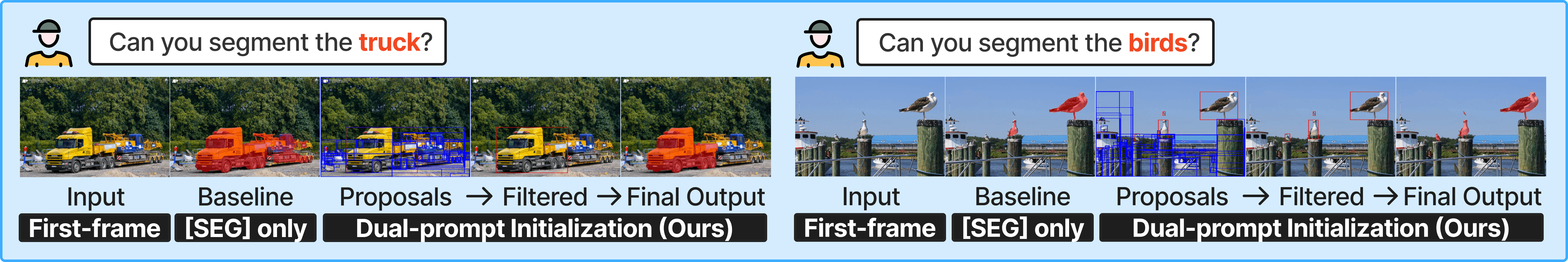}
  \caption{
  \textbf{Illustrative process of the dual-prompt initialization.}
  Given a query (e.g., ``Can you segment the truck?''), our module first generates class-agnostic proposals, which are filtered by the \texttt{[BOX]} prompt and then refined by the \texttt{[SEG]} prompt to produce precise segmentation masks. The dual-prompt approach provides tighter localization and sharper boundaries compared with using \texttt{[SEG]} only.
  }
  \label{fig:dual_prompt_init_example}
\end{figure*}

\paragraph{Illustrative Example.}
As illustrated in Fig.~\ref{fig:dual_prompt_init_example}, combining \texttt{[BOX]} and \texttt{[SEG]} prompts enables coarse-to-fine grounding that improves both spatial localization and mask precision. The \texttt{[BOX]} prompt initializes geometric priors by selecting relevant proposals through the filtration head, while the \texttt{[SEG]} prompt refines these regions with language-conditioned semantics. Compared to using \texttt{[SEG]} alone, the dual-prompt initialization stabilizes first-frame grounding, reduces drift, and yields more precise segmentation.

\subsection{Training Strategy}
\label{subsec:train}

\paragraph{Stage 1: Target-Specific Information Injection.}
After constructing the TSF dataset in Sec.~\ref{sec:tsf}, we train the multimodal model to exploit temporally aligned referent cues.
The goal is to teach the model identity persistence and referential alignment across frames without requiring tracking at inference.
We optimize only the multimodal adapters and lightweight language parameters while keeping the visual backbone and pixel decoders frozen.
Specifically, we update the V$\!\rightarrow$L adapters $(\mathcal{W}_g,\mathcal{W}_h)$, the L$\!\rightarrow$V SEG adapters $\mathcal{W}_s$, and the LoRA parameters~\cite{lora} within the LLM.
This isolates learning to cross-modal pathways and the language backbone, preserving pretrained visual features.

To demonstrate the modularity of our approach, we adopt the training structures of two state-of-the-art video MLLMs~\cite{videoglamm,glus} and extend them with the TSF dataset for temporal referential supervision.
Adapters are optimized with bidirectional alignment between video and language embeddings.
For each video–text pair $(v, q)$, the model encodes global representations $(\mathbf{z}_v,\mathbf{z}_q)$ that include TSF tokens, and the total objective is
\begin{equation}
    \mathcal{L}_{\text{total}} = \mathcal{L}_{\text{CE}} + \mathcal{L}_{\text{mask}},
    \quad
    \mathcal{L}_{\text{mask}} = \mathcal{L}_{\text{BCE}} + \mathcal{L}_{\text{DICE}},
\end{equation}
where $\mathcal{L}_{\text{CE}}$ enforces semantic alignment between vision and language tokens, and $\mathcal{L}_{\text{mask}}$ supervises pixel grounding through binary cross-entropy and Dice losses.
Stage~1 is trained until convergence and then frozen in the next stage.

\paragraph{Stage 2: Bounding Box Prompt Learning.}
This stage focuses on the dual-prompt framework introduced in Sec.~\ref{subsec:dualprompt}, combining an independently trained region proposer with filter head tuning.

\noindent \textit{Proposal Generator Pretraining.}
We first train a class-agnostic region proposer independent of the MLLM training pipeline.
It consists of a Deformable-DETR (D-DETR) \cite{deformable_detr} head operating on frozen Hiera features from SAM2.
Following~\cite{groma}, the proposer is pretrained on a unified corpus of object detection datasets, including COCO~\cite{coco}, Objects365~\cite{objects365}, OpenImages~\cite{openimages}, and V3Det~\cite{v3det}.
All category labels are discarded to enforce class-agnostic localization.
The objective combines three terms:
\begin{equation}
    \mathcal{L}_{\text{prop}} 
    = \mathcal{L}_{\text{obj}} 
    + \lambda_1 \mathcal{L}_{\ell_1} 
    + \lambda_2 \mathcal{L}_{\text{GIoU}},
\end{equation}
where $\mathcal{L}_{\text{obj}}$ is binary cross-entropy for objectness,
$\mathcal{L}_{\ell_1}$ penalizes coordinate errors, and
$\mathcal{L}_{\text{GIoU}}$ enforces geometric consistency.

\noindent \textit{Task Adaptation: Filter-Only Fine-Tuning.}
We adapt the model to task-specific grounding datasets by fine-tuning a lightweight filtration head $\mathcal{F}_{\text{filter}}$, together with the L$\!\rightarrow$V BOX adapter $\mathcal{W}_b$, while keeping all other modules frozen. 
The module $\mathcal{F}_{\text{filter}}$ learns to score and refine region proposals conditioned on textual queries, enabling efficient adaptation while preserving the general objectness learned during pretraining. 
All remaining components, including the Hiera encoder, region proposer, adapters $(\mathcal{W}_g,\mathcal{W}_h,\mathcal{W}_s)$, the LLM backbone, and SAM2, are kept frozen to ensure no catastrophic forgetting.
Training uses datasets with paired textual queries and annotated boxes or masks (converted to boxes at this stage). 
Given a frozen proposal set $\{(b_i,p_i^{\text{det}})\}_{i=1}^{N}$, where $b_i$ is the $i$-th box and $p_i^{\text{det}}$ its objectness score, $\mathcal{F}_{\text{filter}}$ predicts a language-conditioned confidence $s_i\!\in\![0,1]$ and a refined box $\hat{b}_i$. 
Since the SAM2 decoder is frozen, mask supervision does not backpropagate, and optimization is restricted to proposal-level losses:
\begin{equation}
\mathcal{L}_{\text{filter}}
= \lambda_{\text{cls}} \mathcal{L}_{\text{BCE}}
+ \lambda_{\text{box}}
\big(\mathcal{L}_{\ell_1} + \mathcal{L}_{\text{GIoU}}\big),
\end{equation}
where $\mathcal{L}_{\text{BCE}}$ supervises proposal–language alignment, and $\mathcal{L}_{\ell_1}$ and $\mathcal{L}_{\text{GIoU}}$ refine positive boxes. 
Proposals with IoU $>0.5$ are treated as positives and those with IoU $<0.2$ as negatives. 
We set $\lambda_{\text{cls}}\!=\!1.0$ and $\lambda_{\text{box}}\!=\!2.0$ as the default.

\begin{figure*}[!t]
    \centering
    \includegraphics[width=\textwidth]{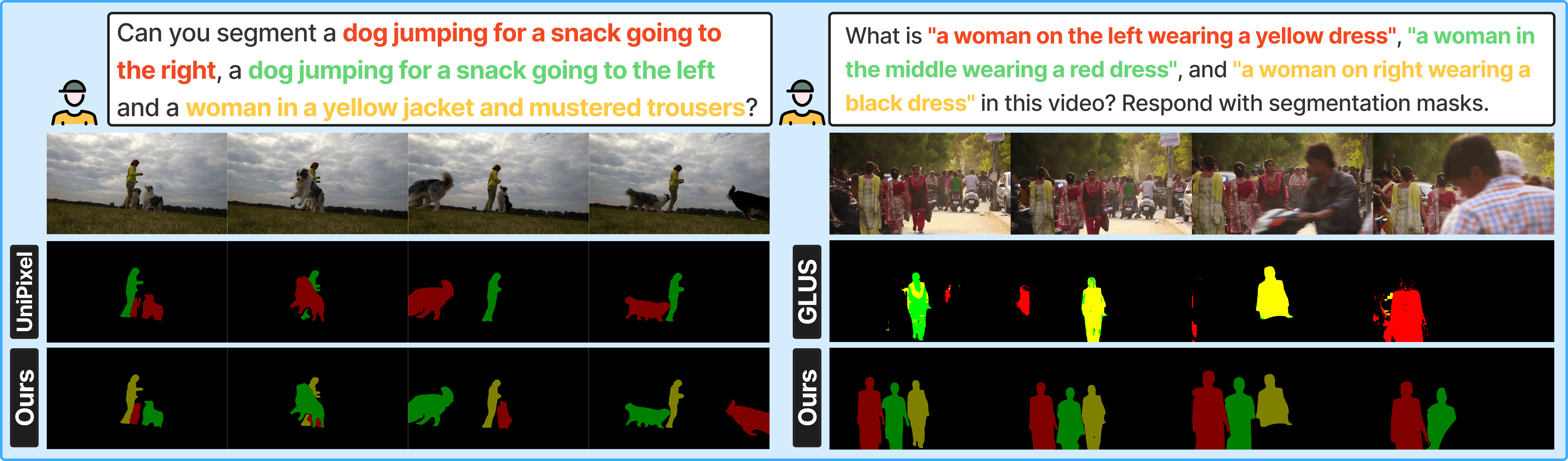}
    \caption{
    \textbf{Qualitative comparison on the Referring Video Object Segmentation (RVOS) task.}
    (Left) \textit{Dog scene:} \textbf{SPARROW} accurately segments “dog … to the right,” “dog … to the left,” and “woman in a yellow jacket” from the first frame, preserving their identities across motion and overlap. 
    (Right) \textit{Crowd scene:} \textbf{SPARROW} cleanly separates “the woman in yellow,” “the woman in red,” and “the woman in black,” maintaining consistent masks and sharp boundaries under occlusion and scale variation. 
    In contrast, UniPixel~\cite{unipixel} and GLUS~\cite{glus} exhibit early inaccuracies, mask swaps, and boundary artifacts across both scenes. 
    The dual-prompt initialization enables precise first-frame grounding and referentially consistent segmentation, producing temporally stable masks throughout the sequence.
    }
\label{fig:qualitative_rvos}
\end{figure*}

\section{Experiments} \label{sec:results}
\paragraph{Setup.}
We build \textbf{SPARROW} upon the baselines \cite{unipixel,videoglamm,glus}, following their released protocols, and evaluate on six datasets spanning three tasks:
(i) MeViS~\cite{mevis} (\textit{val} and \textit{val$^{\text{u}}$}) and (ii) Ref-YouTube-VOS (Ref-YTVOS)~\cite{refervos} and (iii) Ref-DAVIS17~\cite{referdavis} for Referring Video Object Segmentation (RVOS),
(iv) VidSTG~\cite{vidstg} for video visual grounding,
and (v) Video Grounded Conversation Generation (GCG)~\cite{videoglamm}.
Unless stated, we follow each benchmark’s official protocol and report $\mathcal{J}$, $\mathcal{F}$, and $\mathcal{J\&F}$ for RVOS, mIoU for VidSTG, and mIoU/Recall plus METEOR/CIDEr/CLAIR for GCG.
Here, $\mathcal{J}$ measures region similarity (IoU), $\mathcal{F}$ evaluates contour accuracy (boundary quality), and $\mathcal{J\&F}$ denotes their mean, which reflects overall segmentation quality.
We adopt the same prompt templates across baselines and our method. Further details on datasets, evaluation setups, and additional qualitative evaluations can be found in the Appendices 
\ref{app:datasets}, \ref{app:eval_setup}, and \ref{app:qualitative}. 

\subsection{Referring Video Object Segmentation (RVOS)} \label{RVOS}

\paragraph{Protocol.}
The model receives a video and a referring expression and outputs per-frame masks.
We use the template such as: \textit{``What is \{Phrase\} in this video? Respond with segmentation masks.''}
The model responds by producing \texttt{\{[SEG]\,[BOX]\}}, where \texttt{[SEG]} seeds the mask decoder and our additional \texttt{[BOX]} conditions the class-agnostic proposer (Sec.~\ref{subsec:dualprompt}). 


\paragraph{MeViS.}
MeViS evaluates RVOS with motion expressions, emphasizing temporal reasoning and motion-grounded localization. As shown in Table~\ref{tab:mevis}, SPARROW consistently improves performance across all backbones and splits. When integrated into VideoGLaMM, it increases the unseen split (\texttt{val$^{\text{u}}$}) score by \textbf{+8.9} in $\mathcal{J\&F}$, demonstrating stronger compositional generalization to unseen motion–language combinations. Similar improvements are observed on the stronger UniPixel and GLUS backbones, where gains range from \textbf{+0.3} to \textbf{+3.4} across both splits. Across all models, SPARROW achieves the best overall results on MeViS, showing that its temporally supervised training and dual-prompt grounding universally enhance motion-driven segmentation, regardless of backbone strength.

\begin{table}[t!]
\centering
\caption{\textbf{MeViS}~\cite{mevis} results on \texttt{val} and \texttt{val$^{\text{u}}$}, which assess seen and unseen motion-expression compositions, respectively.}
\resizebox{0.99\linewidth}{!}{
\begin{tabular}{l|ccc|ccc}
\specialrule{1pt}{0pt}{0pt}
\multirow{2.6}{*}{\textbf{Method}} 
& \multicolumn{3}{c|}{\textbf{MeViS (\texttt{val})}}
& \multicolumn{3}{c}{\textbf{MeViS (\texttt{val$^{\text{u}}$})}}
\\
\cmidrule(lr){2-4} \cmidrule(lr){5-7}
& $\mathcal{J}$ & $\mathcal{F}$ & $\mathcal{J\&F}$ & $\mathcal{J}$ & $\mathcal{F}$ & $\mathcal{J\&F}$ \\
\specialrule{1pt}{0pt}{0pt}
UniPixel\textsubscript{Neurips'25} ~\cite{unipixel}
& 50.4 & 55.7 & 53.1 
& 56.1 & 63.2 & 59.7 \\
\rowcolor{blue!7.5}
\,+ SPARROW 
& \begin{tabular}[c]{@{}c@{}}
\textbf{51.6}\\ 
\bgain{+1.2}\end{tabular}
& \begin{tabular}[c]{@{}c@{}}\textbf{57.2}\\ \bgain{+1.5}\end{tabular} 
& \begin{tabular}[c]{@{}c@{}}\textbf{54.4}\\ \bgain{+1.3}\end{tabular}
& \begin{tabular}[c]{@{}c@{}}\textbf{57.2}\\ \bgain{+1.1}\end{tabular} 
& \begin{tabular}[c]{@{}c@{}}64.2\\ \bgain{+1.0}\end{tabular} 
& \begin{tabular}[c]{@{}c@{}}60.7\\ \bgain{+1.0}\end{tabular}
\\
\specialrule{1pt}{0pt}{0pt}
GLUS\textsubscript{CVPR'25}~\cite{glus}
& 48.5 & 54.2 & 51.3 
& 56.1 & 63.4 & 59.8 \\
\rowcolor{blue!7.5}
\,+ SPARROW
& \begin{tabular}[c]{@{}c@{}}50.4\\ \bgain{+1.9}\end{tabular}
& \begin{tabular}[c]{@{}c@{}}56.0\\ \bgain{+1.8}\end{tabular}
& \begin{tabular}[c]{@{}c@{}}53.2\\ \bgain{+1.9}\end{tabular}
& \begin{tabular}[c]{@{}c@{}}57.0\\ \bgain{+0.9}\end{tabular}
& \begin{tabular}[c]{@{}c@{}}\textbf{66.8}\\ \bgain{+3.4}\end{tabular}
& \begin{tabular}[c]{@{}c@{}}\textbf{61.9}\\ \bgain{+0.3}\end{tabular} \\
\specialrule{1pt}{0pt}{0pt}
VideoGLaMM\textsubscript{CVPR'24}~\cite{videoglamm}
& 42.1 & 48.2 & 45.2 
& 44.0 & 52.9 & 48.5 \\
\rowcolor{blue!7.5}
\,+ SPARROW 
& \begin{tabular}[c]{@{}c@{}}45.3\\ \bgain{+3.2}\end{tabular}
& \begin{tabular}[c]{@{}c@{}}49.7\\ \bgain{+1.5}\end{tabular}
& \begin{tabular}[c]{@{}c@{}}47.5\\ \bgain{+2.3}\end{tabular}
& \begin{tabular}[c]{@{}c@{}}54.6\\ \bgain{+10.6}\end{tabular}
& \begin{tabular}[c]{@{}c@{}}60.2\\ \bgain{+7.3}\end{tabular}
& \begin{tabular}[c]{@{}c@{}}57.4\\ \bgain{+8.9}\end{tabular} \\
\specialrule{1pt}{0pt}{0pt}
\end{tabular}
}
\label{tab:mevis}
\end{table}
%

\begin{table}[t!]
\centering
\caption{\textbf{Ref-YTVOS}~\cite{refervos} and \textbf{Ref-DAVIS17}~\cite{referdavis} \texttt{(val)} results, where Ref-YTVOS emphasizes large-scale linguistic grounding and Ref-DAVIS17 focuses on fine-grained boundary accuracy.}
\resizebox{0.99\linewidth}{!}{
\begin{tabular}{l|ccc|ccc}
\specialrule{1pt}{0pt}{0pt}
\multirow{2.6}{*}{\textbf{Method}} 
& \multicolumn{3}{c|}{\textbf{Ref-YTVOS}} 
& \multicolumn{3}{c}{\textbf{Ref-DAVIS17}} \\
\cmidrule(lr){2-4} \cmidrule(lr){5-7}
& $\mathcal{J}$ & $\mathcal{F}$ & $\mathcal{J\&F}$ 
& $\mathcal{J}$ & $\mathcal{F}$ & $\mathcal{J\&F}$ \\
\specialrule{1pt}{0pt}{0pt}
UniPixel\textsubscript{Neurips'25} ~\cite{unipixel} 
& 68.6 & 72.3 & 70.5 
& 70.7 & 77.8 & 74.2 \\
\rowcolor{blue!7.5}
\,+ SPARROW 
& \begin{tabular}[c]{@{}c@{}}\textbf{69.0}\\ \bgain{+0.4}\end{tabular}
& \begin{tabular}[c]{@{}c@{}}\textbf{72.4}\\ \bgain{+0.1}\end{tabular}
& \begin{tabular}[c]{@{}c@{}}\textbf{70.7}\\ \bgain{+0.2}\end{tabular}

& \begin{tabular}[c]{@{}c@{}}71.5\\ \bgain{+0.8}\end{tabular}
& \begin{tabular}[c]{@{}c@{}}\textbf{81.2}\\ \bgain{+3.4}\end{tabular}
& \begin{tabular}[c]{@{}c@{}}76.4\\ \bgain{+2.2}\end{tabular}
\\
\specialrule{1pt}{0pt}{0pt}
GLUS\textsubscript{CVPR'25} ~\cite{glus} 
& 65.5 & 69.0 & 67.3 
& 69.7 & 76.1 & 72.9 \\
\rowcolor{blue!7.5}
\,+ SPARROW
& \begin{tabular}[c]{@{}c@{}}67.3\\ \bgain{+1.8}\end{tabular}
& \begin{tabular}[c]{@{}c@{}}70.9\\ \bgain{+1.9}\end{tabular}
& \begin{tabular}[c]{@{}c@{}}69.1\\ \bgain{+1.8}\end{tabular}
& \begin{tabular}[c]{@{}c@{}}70.6\\ \bgain{+0.9}\end{tabular}
& \begin{tabular}[c]{@{}c@{}}80.3\\ \bgain{+4.2}\end{tabular}
& \begin{tabular}[c]{@{}c@{}}75.5\\ \bgain{+2.6}\end{tabular} \\
\specialrule{1pt}{0pt}{0pt}
VideoGLaMM\textsubscript{CVPR'24} ~\cite{videoglamm} 
& 65.4 & 68.2 & 66.8 
& 73.3 & 65.6 & 69.5 \\
\rowcolor{blue!7.5}
\,+ SPARROW 
& \begin{tabular}[c]{@{}c@{}}67.1\\ \bgain{+1.7}\end{tabular}
& \begin{tabular}[c]{@{}c@{}}70.7\\ \bgain{+2.5}\end{tabular}
& \begin{tabular}[c]{@{}c@{}}68.9\\ \bgain{+2.1}\end{tabular}
& \begin{tabular}[c]{@{}c@{}}\textbf{73.4}\\ \bgain{+0.1}\end{tabular}
& \begin{tabular}[c]{@{}c@{}}80.1\\ \bgain{+14.5}\end{tabular}
& \begin{tabular}[c]{@{}c@{}}\textbf{76.8}\\ \bgain{+7.3}\end{tabular} \\
\specialrule{1pt}{0pt}{0pt}
\end{tabular}
}
\label{tab:ytvos_davis}
\vspace{-1em}
\end{table}

\paragraph{Ref-YTVOS \& Ref-DAVIS17.}
Table~\ref{tab:ytvos_davis} summarizes two complementary RVOS regimes. 
Ref-YTVOS focuses on large-scale linguistic diversity with shorter YouTube clips, while Ref-DAVIS17 emphasizes fine-grained contour accuracy and temporal stability across densely annotated sequences. 
SPARROW improves all three backbones on both datasets. 
On Ref-YTVOS, gains are balanced across region overlap ($\mathcal{J}$) and boundary quality ($\mathcal{F}$), with the largest $\mathcal{J\&F}$ increase of about \textbf{+2.1} on VideoGLaMM and consistent gains for GLUS and UniPixel in the \textbf{+0.2}–\textbf{+1.9} range. 
On Ref-DAVIS17, improvements are dominated by $\mathcal{F}$, where boundary accuracy increases by up to \textbf{+14.5} and leads to \textbf{+7.3} in $\mathcal{J\&F}$ on VideoGLaMM, while GLUS and UniPixel also show steady boosts of \textbf{+2.6} and \textbf{+2.2} in $\mathcal{J\&F}$, with all the $\mathcal{F}$ surpassing \textbf{80} score.
Qualitative examples in Fig.~\ref{fig:qualitative_rvos} illustrate these trends, showing that SPARROW delivers precise first-frame masks and maintains consistent boundaries under motion and occlusion. 
Across both datasets, SPARROW achieves the best overall performance across all metrics and backbones, 
demonstrating its robustness in improving temporal stability, spatial precision, and referential consistency under different RVOS settings.
\subsection{Video Visual Grounding (VG)} \label{vg}
\paragraph{Protocol.}
This task aims to localize and segment the region in a video corresponding to a natural-language query, requiring models to align spatial cues with linguistic semantics under temporal variation. 
Using VidSTG~\cite{vidstg}, we query the model with an interrogative caption, for example: \textit{``\{caption\} Please respond with a segmentation mask.''} The model will predict language-conditioned masks for the referred entity, similar to RVOS task (Sec.~\ref{RVOS}).

\begin{figure}[t]
  \centering
  \begin{tikzpicture}
    \begin{axis}[
      ybar,
      width=0.9\linewidth, height=4.2cm,
      bar width=10pt,
      ymin=25, ymax=50,
      ylabel={VidSTG (I-mIoU)},
      symbolic x coords={UniPixel, GLUS, VideoGLaMM},
      xtick=data,
      legend style={at={(0.5,1.05)},anchor=south,legend columns=-1},
      ymajorgrids,
      enlarge x limits=0.2,
      nodes near coords,
      nodes near coords align={vertical},
      nodes near coords style={font=\footnotesize, /pgf/number format/precision=2}
    ]
      \addplot coordinates {(UniPixel,41.25) (GLUS,29.92) (VideoGLaMM,39.66)};
      \addplot coordinates {(UniPixel,46.74) (GLUS,35.17) (VideoGLaMM,45.06)};
      \legend{Baseline,+ SPARROW}
    \end{axis}
  \end{tikzpicture}
  \caption{\textbf{Visual grounding on VidSTG (interrogative).}
  SPARROW boosts mIoU by \textbf{+5.49} on UniPixel (41.25$\rightarrow$46.74), \textbf{+5.25} on GLUS (29.92$\rightarrow$35.17), and \textbf{+5.40} on VideoGLaMM (39.66$\rightarrow$45.06), consistently improving spatial and mask quality.}
  \label{fig:vidstg_bar}
\end{figure}
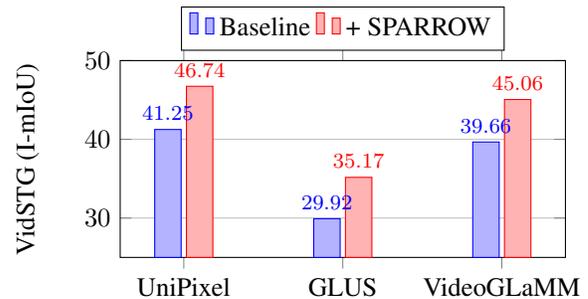

\paragraph{Results.}
Figure~\ref{fig:vidstg_bar} shows that SPARROW achieves around a \textbf{five-point} mIoU improvement across UniPixel, GLUS, and VideoGLaMM, corresponding to roughly \textbf{13–18\%} relative gain. 
The improvements are consistent across all backbones, with the largest relative increase on GLUS and comparable absolute gains on the stronger UniPixel and VideoGLaMM. 
Compared with RVOS, where improvements are split between region overlap and boundary accuracy, VidSTG measures both aspects in a single mIoU score. 
The uniform gains confirm that SPARROW’s dual-prompt design enhances spatial precision and referential grounding in language-driven video understanding. 
Qualitative examples illustrating these effects are provided in Appendix~\ref{app:qualitative}.

\begin{table}[t]
\centering
\caption{\textbf{VideoGCG benchmark}~\cite{videoglamm}. We report mask quality (mIoU, Recall) and caption quality (METEOR, CIDEr, CLAIR).}
\resizebox{0.99\columnwidth}{!}{
\begin{tabular}{l|ccccc}
\specialrule{1pt}{0pt}{0pt}
\textbf{Model} & \textbf{mIoU} & \textbf{Recall} & \textbf{METEOR} & \textbf{CIDEr} & \textbf{CLAIR} \\
\specialrule{1pt}{0pt}{0pt}
UniPixel\textsubscript{Neurips'25} ~\cite{unipixel} & 52.0 & 0.311 & 0.095 & 0.55 & 26.0 \\
\rowcolor{blue!7.5}
\,+ SPARROW 
& \begin{tabular}[c]{@{}c@{}}\textbf{54.5}\\ \bgain{+2.5}\end{tabular}
& \begin{tabular}[c]{@{}c@{}}\textbf{0.325}\\ \bgain{+0.01}\end{tabular}
& \begin{tabular}[c]{@{}c@{}}\textbf{0.103}\\ \bgain{+0.01}\end{tabular}
& \begin{tabular}[c]{@{}c@{}}\textbf{0.58}\\ \bgain{+0.03}\end{tabular}
& \begin{tabular}[c]{@{}c@{}}\textbf{29.4}\\ \bgain{+3.4}\end{tabular}
\\
\specialrule{1pt}{0pt}{0pt}
GLUS\textsubscript{CVPR'25} ~\cite{glus} & 45.86 & 0.245 & -- & -- & -- \\
\rowcolor{blue!7.5}
\,+ SPARROW 
& \begin{tabular}[c]{@{}c@{}}\textbf{47.91}\\ \bgain{+2.05}\end{tabular}
& \begin{tabular}[c]{@{}c@{}}\textbf{0.265}\\ \bgain{+0.02}\end{tabular}
& -- & -- & -- \\
\specialrule{1pt}{0pt}{0pt}
VideoGLaMM\textsubscript{CVPR'24} ~\cite{videoglamm} & 62.34 & 0.375 & 0.103 & 0.59 & 28.2 \\
\rowcolor{blue!7.5}
\,+ SPARROW & 
\begin{tabular}[c]{@{}c@{}}\textbf{65.59}\\ \bgain{+3.25}\end{tabular} &
\begin{tabular}[c]{@{}c@{}}\textbf{0.383}\\ \bgain{+0.01}\end{tabular} &
\begin{tabular}[c]{@{}c@{}}\textbf{0.112}\\ \bgain{+0.01}\end{tabular} &
\begin{tabular}[c]{@{}c@{}}\textbf{0.63}\\ \bgain{+0.04}\end{tabular} &
\begin{tabular}[c]{@{}c@{}}\textbf{33.6}\\ \bgain{+5.4}\end{tabular} \\
\specialrule{1pt}{0pt}{0pt}
\end{tabular}
}
\footnotesize{\textit{Note:} GLUS emits only \texttt{[SEG]} tokens (no free text); language metrics are therefore not applicable.}
\label{tab:videogcg}
\end{table}

\subsection{Video Grounded Conversation Generation} \label{GCG}
\paragraph{Protocol.}
This task requires a model to produce a natural-language description while grounding key phrases with pixel-level masks. 
We follow the VideoGLAMM~\cite{videoglamm} setup and use a fixed prompt: 
\textit{``Could you please give me a detailed description of the video? Please respond with interleaved segmentation masks for the corresponding parts of the answer.''} 
For UniPixel and GLUS, which do not natively support this task, we fine-tuned them on the VideoGCG train dataset to enable interleaved text–mask generation.

\paragraph{Results.}
Table~\ref{tab:videogcg} shows that SPARROW consistently improves mask quality by \textbf{+2–3.25} points in mIoU and slightly increases Recall across all backbones. 
For models that generate text, language metrics also improve, with CLAIR showing the most significant gain (e.g., \textbf{+5.4} on VideoGLaMM), indicating better alignment between generated phrases and their corresponding visual regions. 
These results extend the trends observed in RVOS and VG tasks: TSF pretraining and dual-prompt design yield cleaner, more accurate masks and richer, more specific textual descriptions. 
Qualitative examples in Fig.~\ref{fig:gcg} illustrate these improvements with additional samples in Appendix~\ref{app:qualitative}.




\begin{figure*}[t]
  \centering
  \includegraphics[width=\textwidth]{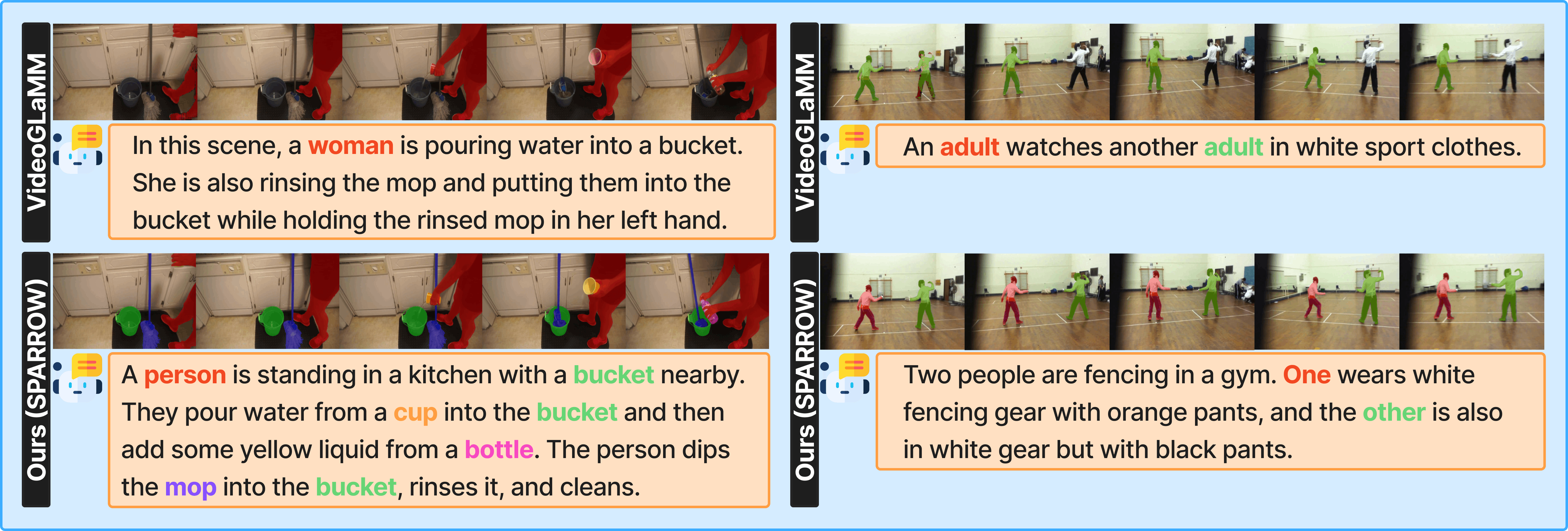}
  \vspace{-2em}
    \caption{
    \textbf{Qualitative comparison on the Grounded Conversation Generation (GCG) task.}
    (Left) Kitchen scene: \textbf{SPARROW} produces more detailed and semantically consistent descriptions (e.g., distinguishing \textit{cup}, \textit{bottle}, and \textit{mop}) compared to VideoGLaMM’s generic output.
    (Right) Fencing scene: \textbf{SPARROW} generates context-rich phrases differentiating individuals by appearance and action, while VideoGLaMM produces shorter, less informative text.
    These examples highlight that \textbf{SPARROW} not only achieves stronger temporal coherence and object grounding, but also narrative completeness across complex scenes.
    }
\label{fig:gcg}
\end{figure*}




\subsection{Ablation Study} \label{Ablation-Study}
Ablations are performed on Ref-DAVIS17 (val) with VideoGLaMM~\cite{videoglamm} as the baseline trained on our corpus.

\paragraph{TSF pretraining and box prompting.}
Table~\ref{tab:tsf_box_matrix} summarizes TSF usage modes (none / train only / train+inference) and the effect of enabling \texttt{[BOX]} prompting. 
Training with TSF (no test-time TSF) improves $\mathcal{J\&F}$ by \textbf{+2.9} over the baseline (69.5$\rightarrow$72.4), showing that pseudo-tracked cues teach identity persistence even when removed at inference.
Activating \texttt{[BOX]} adds \textbf{+3.0$\sim$+8.2} depending on TSF mode, reducing drift via geometric priors. 
The best score (77.7) occurs when both TSF and \texttt{[BOX]} are active at train and test, but the default configuration (TSF at train only, \texttt{[BOX]} on) already captures most gains without runtime tracking. Further trade-off analysis in Appendix~\ref{app:tsf_tradeoff}.

\begin{table}[!t]
\centering
\caption{\textbf{TSF pretraining and \texttt{[BOX]} prompting} on Ref-DAVIS17 (val). Values are $\mathcal{J\&F}$.}
\setlength{\tabcolsep}{17pt}
\scalebox{0.8}[0.8]{
\begin{tabular}{l|cc}
\specialrule{1pt}{0pt}{0pt}
\textbf{TSF usage} & \textbf{BOX OFF} & \textbf{BOX ON} \\
\specialrule{1pt}{0pt}{0pt}
No TSF   & 69.5 \footnotesize{(baseline)}      & 72.5 \textcolor{red}{\footnotesize(\textbf{+3.0})} \\
Train only (our default) & 72.4 \textcolor{red}{\footnotesize(\textbf{+2.9})} & 76.8 \textcolor{red}{\footnotesize(\textbf{+7.3})} \\
\rowcolor{blue!7.5}
Train + Inference & 75.3 \textcolor{red}{\footnotesize(\textbf{+5.8})} & \textbf{77.7} \textcolor{red}{\footnotesize(\textbf{+8.2})} \\
\specialrule{1pt}{0pt}{0pt}
\end{tabular}
}
\label{tab:tsf_box_matrix}
\end{table}

\paragraph{Selection supervision and prompt composition.}
Table~\ref{tab:selection_prompt} compares (i) supervision tokens for the Filtration Head, (ii) selection mechanisms, and (iii) inference prompt composition. 
Using \texttt{[SEG]} features for supervision is less effective than \texttt{[BOX]} (\textbf{70.6} vs.\ \textbf{72.5}, $-1.9$), showing that spatially explicit cues are essential for proposal ranking. 
The detector-side Filtration Head also outperforms MLLM-side selection (\textbf{72.5} vs.\ \textbf{70.4}, $+2.1$). 
At inference, single-token prompts perform worse (\texttt{[SEG]}: \textbf{69.5}, \texttt{[BOX]}: \textbf{68.2}), whereas combining \texttt{[BOX]}+\texttt{[SEG]} reaches \textbf{72.5} (\textbf{+3.0} over the best single), confirming the benefit of dual prompt.


\begin{table}[!t]
\centering
\caption{\textbf{Selection supervision and prompt composition} on Ref-DAVIS17 (val). Values are $\mathcal{J\&F}$.}
\setlength{\tabcolsep}{20pt}
\scalebox{0.8}[0.8]{
\begin{tabular}{l|c}
\specialrule{1pt}{0pt}{0pt}
\multicolumn{2}{l}{\textbf{(A) Filtration Head supervision token}} \\
\specialrule{.6pt}{0pt}{0pt}
\([{\rm SEG}]\) features $\rightarrow$ Filtration Head & 70.6 \\
\rowcolor{blue!7.5}
\([{\rm BOX}]\) features $\rightarrow$ Filtration Head (ours) & \textbf{72.5} \textcolor{red}{\footnotesize(\textbf{+1.9})} \\
\specialrule{.8pt}{0pt}{0pt}
\multicolumn{2}{l}{\textbf{(B) Proposal selection mechanism}} \\
\specialrule{.6pt}{0pt}{0pt}
MLLM filtration (language-only) & 70.4 \\
\rowcolor{blue!7.5}
Filtration Head (ours) & \textbf{72.5} \textcolor{red}{\footnotesize(\textbf{+2.1})} \\
\specialrule{.8pt}{0pt}{0pt}
\multicolumn{2}{l}{\textbf{(C) Inference prompt composition}} \\
\specialrule{.6pt}{0pt}{0pt}
\([{\rm SEG}]\) only & 69.5 \\
\([{\rm BOX}]\) only & 68.2 \\
\rowcolor{blue!7.5}
\([{\rm BOX}]\)+\([{\rm SEG}]\) (ours) & \textbf{72.5} \textcolor{red}{\footnotesize(\textbf{+3.0})} \\
\specialrule{1pt}{0pt}{0pt}
\end{tabular}
}
\label{tab:selection_prompt}
\end{table}



\paragraph{TSF, proposal head design, and cost analysis.}
Additional analyses of TSF overhead, inference guidance, robustness, proposal head variants, and compute/inference 
overhead are provided in Appendices~\ref{app:tsf}, \ref{app:proposal_head}, and \ref{app:compute}.

\paragraph{Discussion and Limitation.}
SPARROW improves temporal consistency and spatial precision but inherits limitations of proposal-based detection. Its performance depends on proposal recall: missed small, occluded, or unseen objects cannot be recovered by downstream \texttt{[BOX]} or \texttt{[SEG]} refinement. Early \texttt{[BOX]} errors may also accumulate in long sequences, although dual prompting mitigates drift compared to \texttt{[SEG]} alone. Moreover, TSF relies on pseudo-tracks from GroundingDINO and CLDTracker; severe tracking noise or ID switches can bias temporal learning, though robustness to such corruption is demonstrated in Appendix~\ref{app:robustness}. Future work includes higher-recall proposals, correction mechanisms, and stronger supervision.

\section{Conclusion}
We presented \textbf{SPARROW}, a pixel-grounded video MLLM that unifies spatial precision with temporally consistent reference tracking. SPARROW introduces Target-Specific Tracked Features (TSF) for temporally aligned supervision and a dual-prompt design combining \texttt{[BOX]} geometric priors with \texttt{[SEG]} semantics through a class-agnostic SAM2-based proposer. We curate a large referential dataset (30{,}646 videos, 45{,}231 Q\&A pairs) to enable scalable TSF training. Its modular design integrates seamlessly with
baselines
without modifying the backbone, yielding consistent gains across 
benchmarks. Ablations confirm that TSF and joint \texttt{[BOX]}+\texttt{[SEG]} prompting stabilize early grounding and reduce temporal drift while preserving flexibility.

\section*{Acknowledgments}
This research was funded by Khalifa University of Science and Technology through the Faculty Start-Ups under Project ID: KU-INT-FSU-2005-8474000775.

{
    \small
    \bibliographystyle{ieeenat_fullname}
    \bibliography{main}
}


\clearpage
\appendix
\clearpage
\setcounter{page}{1}
\setcounter{section}{0}
\maketitlesupplementary
\renewcommand{\thesection}{\Alph{section}}

\begin{itemize}
    \item TSF Datasets (\autoref{app:datasets})
    \item Details on Evaluation Setup (\autoref{app:eval_setup})
    \item Qualitative Analysis (\autoref{app:qualitative})
    \item TSF Usage, Tradeoff, and Design (\autoref{app:tsf})
    \item Comparison of Proposal Head Variants (\autoref{app:proposal_head})
    \item Compute, Training, and Overhead (\autoref{app:compute})
\end{itemize}

\section{TSF Datasets} \label{app:datasets}

\noindent To supervise TSF and dual-prompt grounding at scale, we curate a unified video benchmark tailored for referential understanding and fine-grained segmentation. 
We aggregate seven publicly available datasets, HC-STVG \cite{hcstvg}, VID-Sentence \cite{vidsentence}, A2D Sentences \cite{a2d}, LaSOT \cite{lasot}, MeViS \cite{mevis}, GOT-10k \cite{got10k}, and Ref-SAV \cite{sa2va,sam2}, and process them with a common offline pipeline, yielding a corpus of 30,646 video sequences and 45,231 question--answer pairs. 
In all cases, we standardize annotations into (video, referring text, trajectory, mask) tuples and precompute TSF tokens for SPARROW.

\noindent \textbf{HC-STVG.} \cite{hcstvg} contains movie clips with tracking sequences and natural language descriptions of a person over specific temporal intervals. We use the official training split ($\sim$10K clips) for TSF pretraining and the validation split for evaluation. The original tracking annotations are often noisy, so we regenerate trajectories using GroundingDINO \cite{groundingdino} plus CLDTracker \cite{cldtracker}. 
Specifically, we:
(i) detect the referent in annotated key frames with GroundingDINO,
(ii) track the detected region across the whole clip with CLDTracker to obtain dense trajectories, and
(iii) compare regenerated boxes with the HC-STVG ground truth, discarding frames whose IoU falls below a threshold.
The resulting cleaned tracks are then fed to SAM2 \cite{sam2} to obtain dense masks and to our TSF pipeline (Sec. \ref{sec:tsf}) to generate compact tracked tokens.

\noindent \textbf{A2D Sentences.} \cite{a2d} provides short tracking sequences (often only a few frames) and textual descriptions for multiple actors in each clip. To expose TSF to longer temporal context, we start from the annotated boxes and re-track the referent using CLDTracker, extending trajectories to up to 20 frames where possible. As in HC-STVG, we filter low-IoU frames, convert masks from SAM2 into bounding boxes for box-level supervision, and keep the dense masks for segmentation and TSF supervision.

\noindent \textbf{LaSOT.} \cite{lasot} offers long-term tracking sequences with a single caption describing the target object across the entire video. Since captions are generic and sequences are long, we sub-sample each video into three non-overlapping 10-second segments (fixed-length frame windows) anchored around the target trajectory. Each segment inherits the original caption but is paired with the segment-specific trajectory and masks (obtained via GroundingDINO + CLDTracker + SAM2). This turns each long sequence into several shorter, more focused referential clips.

\noindent \textbf{MeViS.} \cite{mevis} provides videos with dense referring segmentation masks and language expressions. We directly use the official splits, convert the per-frame masks into bounding boxes for proposal-level supervision, and keep the original masks for TSF and dual-prompt training. Since trajectories are already stable, we only run CLDTracker when annotations are sparse or missing in intermediate frames, ensuring temporally continuous supervision.

\noindent \textbf{VID-Sentence.} \cite{vidsentence} contains videos with sentence-level descriptions and sparse annotations. We keep the original temporal spans and regenerate trajectories from a single annotated frame per instance using GroundingDINO and CLDTracker. SAM2 is then applied to produce per-frame masks consistent with the tracked boxes. No additional temporal cropping is used beyond trimming to the annotated span.

\noindent \textbf{GOT-10k.} \cite{got10k} is a tracking benchmark with category, motion, and attribute labels for each tracked object. Following prior work, we convert these labels into natural language descriptions by concatenating them into short sentences (e.g., \textit{``a bear is slowly walking''}). The provided trajectories are treated as initial boxes; we run SAM2 to obtain dense masks and apply our TSF pipeline to select a compact subset of representative regions via K-means clustering in the joint visual--spatial feature space.

\noindent \textbf{Ref-SAV.} \cite{sa2va,sam2} offers long videos with referring expressions and high-quality SAM2-based masks. We treat Ref-SAV as a high-quality source of dense referential supervision: masks are converted into bounding boxes, trajectories are inferred from mask centroids, and TSF tokens are generated from cropped regions without additional tracking.

\noindent \textbf{Offline TSF Pipeline.}
For all datasets, we run a unified offline pipeline that decouples heavy computation from training. Given a video and its referring expression, we:
(i) detect the referent in one or more key frames with GroundingDINO,
(ii) track detections through the clip with CLDTracker to obtain dense trajectories,
(iii) crop tracked regions and encode them with the spatial encoder,
(iv) apply K-means clustering to select $K$ diverse appearances per trajectory, and
(v) project the selected features into the LLM space as TSF tokens $Z_{\text{TSF}}$ (Sec. \ref{sec:tsf}).
Optionally, we invoke SAM2 on the tracked boxes to generate per-frame masks, which are stored alongside trajectories for segmentation supervision.

\noindent \textbf{Question--Answer Construction.}
To convert the curated corpus into training samples suitable for SPARROW, we transform each (video, trajectory, mask, description) tuple into one or more question--answer pairs. We use a small pool of hand-crafted referential templates, such as
\texttt{``What is the <region> doing during this video?''}.
For all datasets we use the caption or description directly as the answer, without modification.
During training, we randomly sample a referential template and fill in the region placeholder, producing diverse yet standardized QA prompts. The final TSF dataset contains 30{,}646 video clips and 45{,}231 question--answer pairs.

\begin{table}[!h]
\centering
\caption{Curated datasets used to construct our TSF dataset.}
\vspace{-1em}
\label{tab:tsf_dataset}
\begin{tabular}{@{}lcc@{}}

\specialrule{1pt}{0pt}{0pt}

\textbf{Dataset} & \textbf{Video Clips} & \textbf{Q\&A Pairs} \\

\specialrule{1pt}{0pt}{0pt}

HC-STVG \cite{hcstvg}          & 10105 & 10105 \\
MeViS \cite{mevis}             & 1644  & 4489  \\
A2D Sentences \cite{a2d}       & 2508  & 5359  \\
LaSOT \cite{lasot}             & 2640  & 7920  \\
VID-Sentence \cite{vidsentence}& 4045  & 7654  \\
GOT-10k \cite{got10k}          & 8195  & 8195  \\
Ref-SAV \cite{sa2va,sam2}      & 1509 & 1509   \\

\specialrule{1pt}{0pt}{0pt}
TSF Dataset            & \textbf{30646} & \textbf{45231} \\
\specialrule{1pt}{0pt}{0pt}
\end{tabular}
\end{table}

\section{Evaluation Setup} \label{app:eval_setup}


\noindent \textbf{VideoGLaMM.}
VideoGLaMM \cite{videoglamm} is a video-centric MLLM designed for dense pixel-level grounding. 
It employs a dual spatio-temporal vision encoder consisting of (i) a CLIP ViT-L/14 image backbone for high-resolution spatial features and (ii) an InternVideo2-based temporal encoder for long-range motion modeling. Both streams are projected into the LLM token space via learnable V$\rightarrow$L adapters and fused with text tokens in a Phi-3 Mini LLM (LoRA-tuned). Grounding is triggered using a special \texttt{<SEG>} token. For mask prediction, VideoGLaMM uses a SAM2-based spatio-temporal decoder that consumes L$\rightarrow$V-transformed LLM embeddings along with multi-scale features from a frozen ViT encoder. Training is end-to-end, combining LLM cross-entropy on grounded captions with IoU-based mask losses.

\noindent \textbf{GLUS.}
GLUS \cite{glus} is an MLLM-based RVOS framework that integrates global and local temporal reasoning. Built on LISA-7B with a SAM2 mask decoder, GLUS divides the video into context frames (uniformly sampled to capture global semantics) and query frames (short temporal windows used for mask prediction). The LLM autoregressively outputs frame-wise \texttt{[SEG]} tokens conditioned on the text query, context frames, and prior query frames. These tokens are decoded into masks using a SAM2 decoder enhanced with an end-to-end learnable memory bank, enabling long-term temporal modeling without an external VOS propagator. GLUS further improves language–object alignment via an object-level contrastive loss over \texttt{[SEG]} tokens and employs a self-refined key-frame selector trained from its own pseudo-IoU confidence.

\noindent \textbf{UniPixel.}
UniPixel~\cite{unipixel} is a unified pixel-grounded MLLM that couples Qwen2.5-VL with a ViT-based visual encoder and a SAM2.1 segmentation head. It encodes visual prompts (points, boxes, masks) into single tokens via Fourier positional encodings combined with temporal embeddings, allowing the LLM to reason over structured visual cues. A central component is the object memory bank, a hashmap storing spatio-temporal object masks. Upon encountering \texttt{<REF>} tokens, the model performs memory pre-fill, segmenting all referenced objects and writing them to memory. During subsequent turns, \texttt{<MEM>} tokens inject object-level features back into the LLM, enabling robust multi-turn and multi-object reasoning. Segmentation is produced via a SAM2 decoder conditioned on downsampled \texttt{<SEG>} token embeddings. UniPixel is trained through a three-stage alignment pipeline over 851k regional captioning, 87k referring segmentation, and 1M mixed pixel-level reasoning samples, using a combination of LM loss, focal/dice loss, MAE, and objectness supervision.


\begin{figure*}[t!]
\centering
\includegraphics[width=0.99\textwidth]{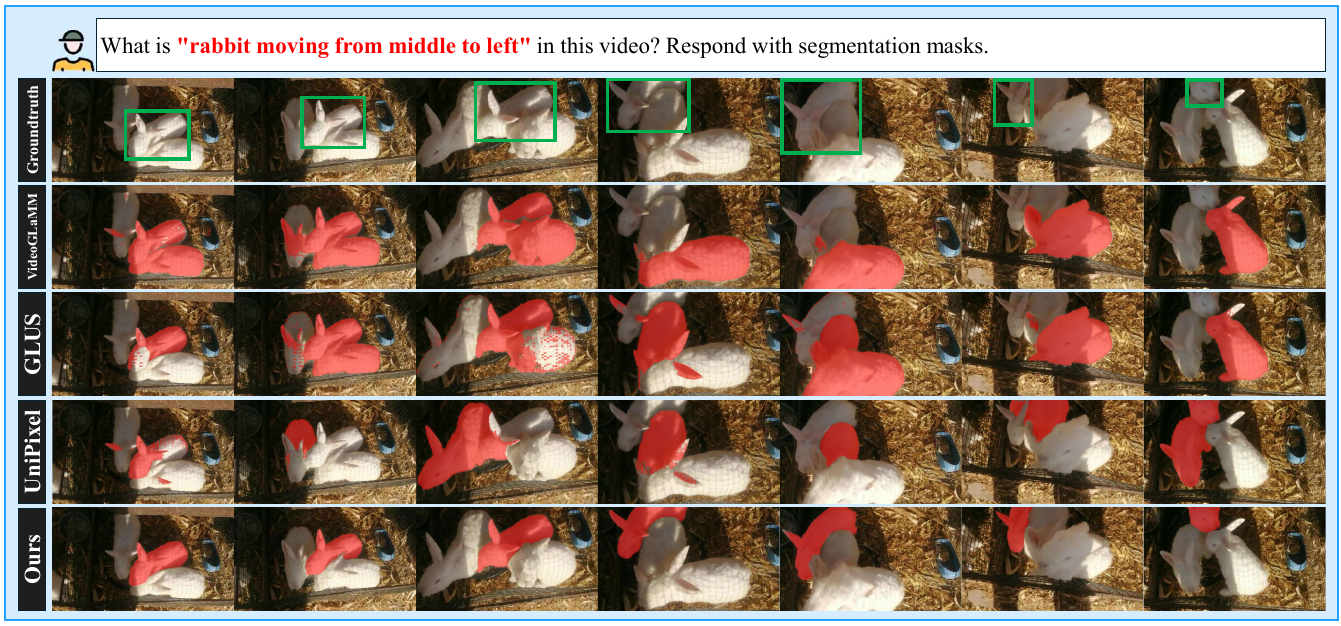}
\caption{
\textbf{Qualitative comparison on the MeViS RVOS task.}
Given the motion-centric query “rabbit moving from middle to left,” our method correctly grounds the intended rabbit from the first frame and maintains its identity throughout the entire sequence, despite the presence of multiple appearance-similar distractors. It produces temporally stable masks with clean boundaries and consistent spatial localization even under close interactions between the rabbits.
In contrast, VideoGLaMM, UniPixel, and GLUS frequently drift between rabbits, mix identities, or generate unstable masks across frames.
The strong referential grounding of our approach enables precise, distraction-robust tracking and segmentation of the queried rabbit across all frames.
}
\label{fig:mevis1}
\end{figure*}
\begin{figure*}[t]
\centering
\includegraphics[width=0.99\textwidth]{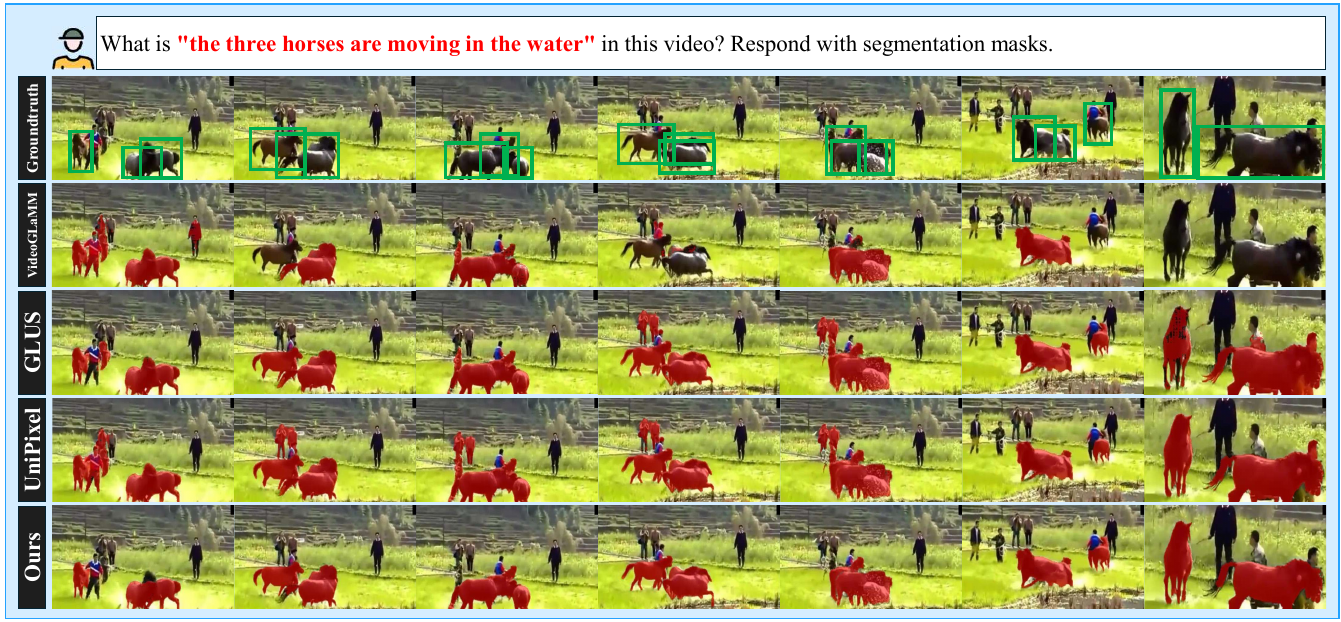}
\caption{
\textbf{Qualitative comparison on the MeViS RVOS task.}
For the query “the three horses are moving in the water,” our method accurately grounds all three horses from the first frame and preserves their individual identities throughout the entire sequence. It maintains clean instance separation, stable temporal masks, and sharp boundaries even as the horses move closely, interact, and occlude one another.
In contrast, VideoGLaMM, UniPixel, and GLUS frequently merge instances, lose one or more horses across frames, or produce inconsistent and drifting masks.
The strong multi-instance grounding of our approach enables reliable segmentation and tracking of all three horses as distinct, temporally consistent targets.
}
\label{fig:mevis2}
\end{figure*}
\begin{figure*}[!t]
\centering
\includegraphics[width=0.99\textwidth]{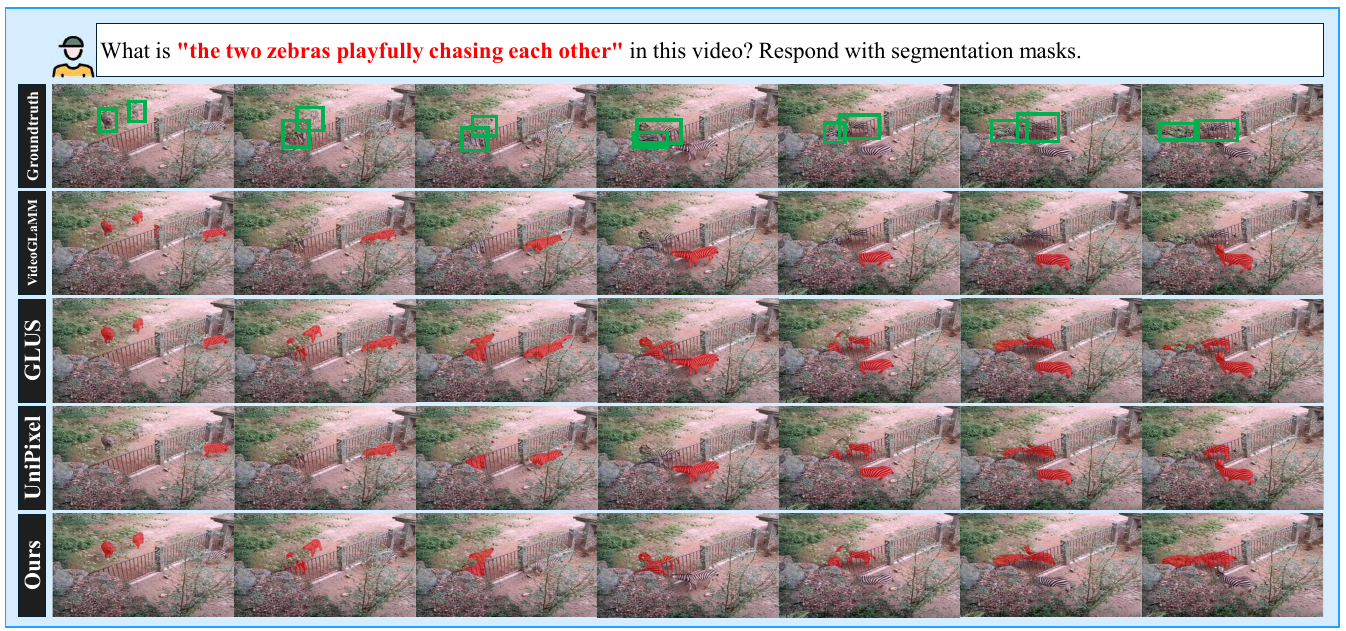}
\caption{
\textbf{Qualitative comparison on the MeViS RVOS task.}
For the query “the two zebras playfully chasing each other,” our method accurately grounds both zebras from the first frame and preserves their identities throughout the sequence, despite rapid motion, tight interaction, and repeated occlusions. It maintains clear instance separation, stable temporal masks, and precise spatial localization under fast, dynamic behavior.
In contrast, VideoGLaMM, UniPixel, and GLUS frequently merge the two zebras, lose one during high-motion frames, or produce unstable and drifting masks.
Our approach delivers robust multi-instance grounding and consistent segmentation of both zebras as distinct targets across the entire video.
}
\label{fig:mevis3}
\end{figure*}

\section{Qualitative Analysis} \label{app:qualitative}

\begin{figure*}[t]
\centering
\includegraphics[width=0.99\textwidth]{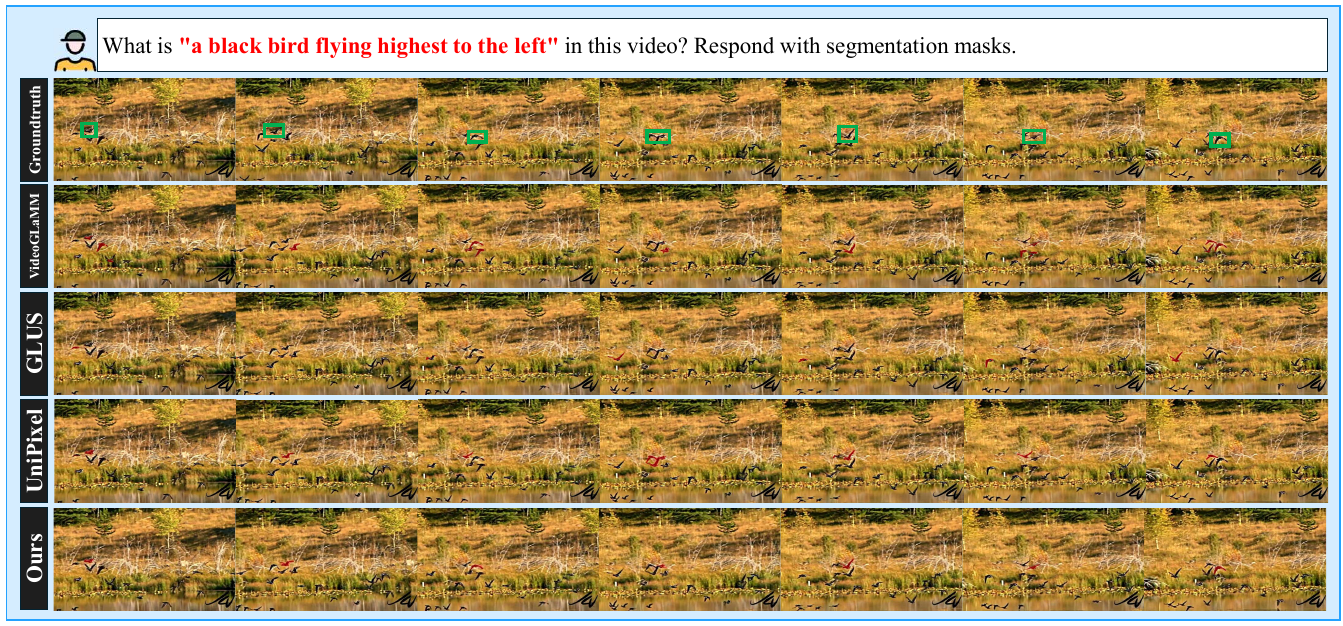}
\caption{
\textbf{Qualitative comparison on the Ref-YTVOS RVOS task.}
For the query “a black bird flying highest to the left,” our method reliably grounds the correct bird from the first frame and maintains accurate tracking throughout the sequence, despite the densely packed flock and rapid aerial motion. It preserves clean spatial localization and stable masks even when multiple birds share nearly identical appearances and flight patterns.
In contrast, VideoGLaMM, GLUS, and UniPixel frequently drift to neighboring birds, lose the target under fast movement, or produce inconsistent and flickering masks.
Our approach demonstrates strong referential grounding and robust temporal consistency in challenging multi-instance, motion-heavy scenarios.
}
\label{fig:ytvos_1}
\end{figure*}

\subsection{Referring Video Object Segmentation (RVOS)}

\noindent \textbf{MeViS.} Fig. \ref{fig:mevis1} shows a case where the motion-centric query ``rabbit moving from middle to left’’ requires distinguishing between two nearly identical rabbits in the same scene. VideoGLaMM, GLUS, and UniPixel often drift onto the stationary rabbit or partially fuse both instances, indicating difficulty maintaining identity under appearance similarity and occlusion. Our SPARROW isolates the correct rabbit throughout the sequence, preserving mask continuity even when the target moves through clutter. This demonstrates the benefit of our dual-prompt grounding and motion-aware temporal cues for queries that hinge on fine motion differences.

Fig. \ref{fig:mevis2} presents a multi-object scenario with the expression “the three horses are moving in the water.” The methods compared struggle to maintain three distinct trajectories: VideoGLaMM and GLUS frequently merge adjacent horses, while UniPixel intermittently loses instances during overlap or viewpoint changes. Our SPARROW keeps all three horses separated and stable across the entire clip, even during heavy interaction. This indicates that our grounding mechanism handles coordinated multi-object motion more reliably than prior pixel-grounding models.

Fig. \ref{fig:mevis3} highlights a more dynamic interaction: “the two zebras playfully chasing each other.” The fast motion, repeated occlusions, and near-identical appearance make this a failure point for existing methods. VideoGLaMM tends to focus on a single zebra; UniPixel shows unstable boundaries and occasional identity swaps; GLUS often collapses both zebras into one mask. Our SPARROW keeps the two zebras separated throughout, even as they cross paths. This example underscores the advantage of our instance-aware temporal modeling in scenes where motion and appearance cues are both ambiguous.

\begin{figure*}[t]
\centering
\includegraphics[width=0.99\textwidth]{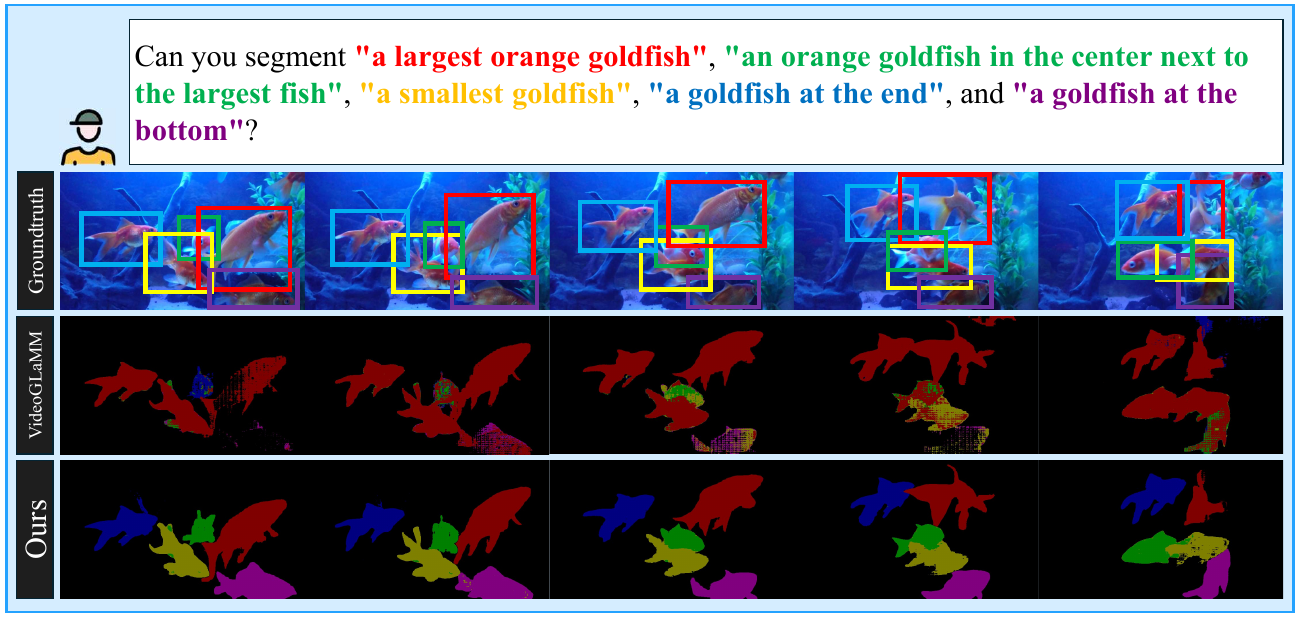}
\caption{
\textbf{Qualitative comparison on the Ref-DAVIS17 RVOS task.}
Given multiple appearance-based prompts targeting distinct goldfish (“largest,” “smallest,” “center,” “end,” “bottom”), our method accurately grounds each described instance from the first frame and maintains clear separation throughout the sequence. It preserves stable masks, fine-grained localization, and consistent identities even when the fish exhibit similar colors, overlapping motion, and close interactions.
In contrast, VideoGLaMM frequently merges instances, confuses similarly colored fish, or produces unstable and drifting segmentations.
Our approach demonstrates strong referential grounding under simultaneous multi-query conditions, delivering reliable instance-specific masks across the entire sequence.
}
\label{fig:davis_videoglamm}
\end{figure*}

\noindent \textbf{Ref-YTVOS.}
Fig. \ref{fig:ytvos_1} features a dense flock scenario with the query \textit{“a black bird flying highest to the left.”} The scene contains many visually similar birds with overlapping motion and strong blur, making appearance-based discrimination unreliable. VideoGLaMM and GLUS frequently jump to nearby birds with similar trajectories, and their masks destabilize as the flock spreads. UniPixel occasionally identifies the correct target but struggles to maintain identity during fast movement, often switching to birds at comparable heights.
SPARROW consistently isolates the correct bird across the entire sequence. It maintains identity despite intersecting flight paths and brief occlusions, showing that the model can track a specific target even when spatial cues are subtle and temporal ambiguity is high. This example highlights SPARROW’s advantage in large, multi-instance scenes where appearance similarity and rapid motion typically cause grounding failures.

\begin{figure*}[t]
\centering
\includegraphics[width=0.99\textwidth]{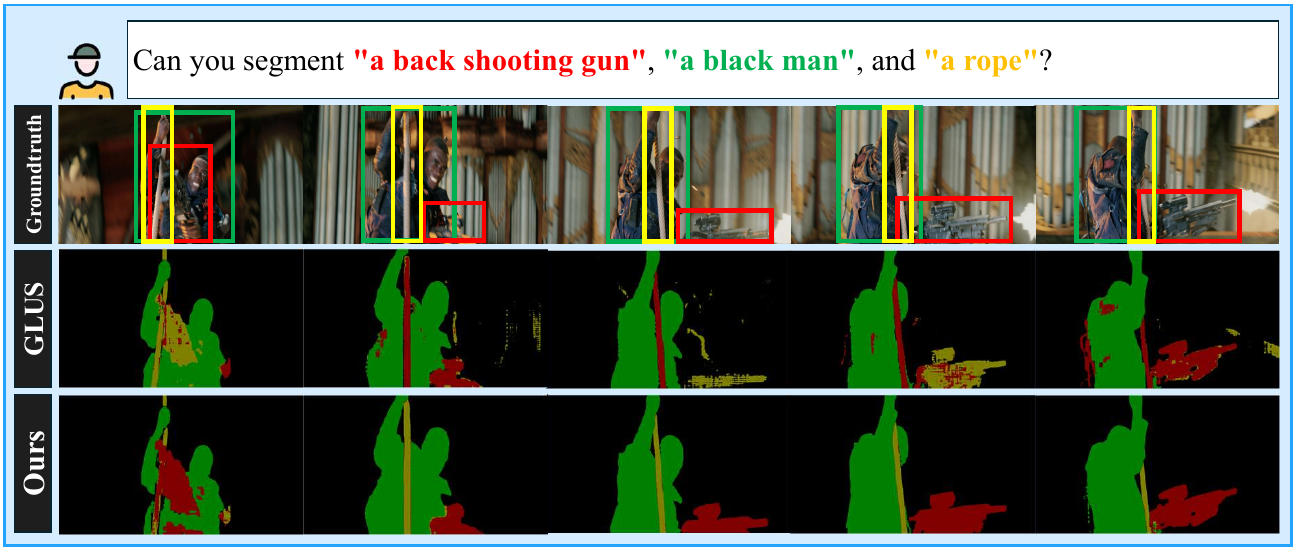}
\caption{
\textbf{Qualitative comparison on the Ref-DAVIS17 RVOS task.}
For the queries “a back shooting gun,” “a black man,” and “a rope,” our method cleanly grounds all three described targets from the first frame and maintains precise, instance-specific masks throughout the sequence. It preserves accurate boundaries and stable localization even on thin or small objects such as the gun barrel and rope.
In contrast, GLUS produces unstable, overlapping masks, frequently blending instances or losing fine structural details under motion and occlusion.
Our approach demonstrates stronger fine-grained appearance grounding and consistent multi-object separation across all frames.
}
\label{fig:davis_glus}
\end{figure*}

\begin{figure*}[t]
\centering
\includegraphics[width=0.99\textwidth]{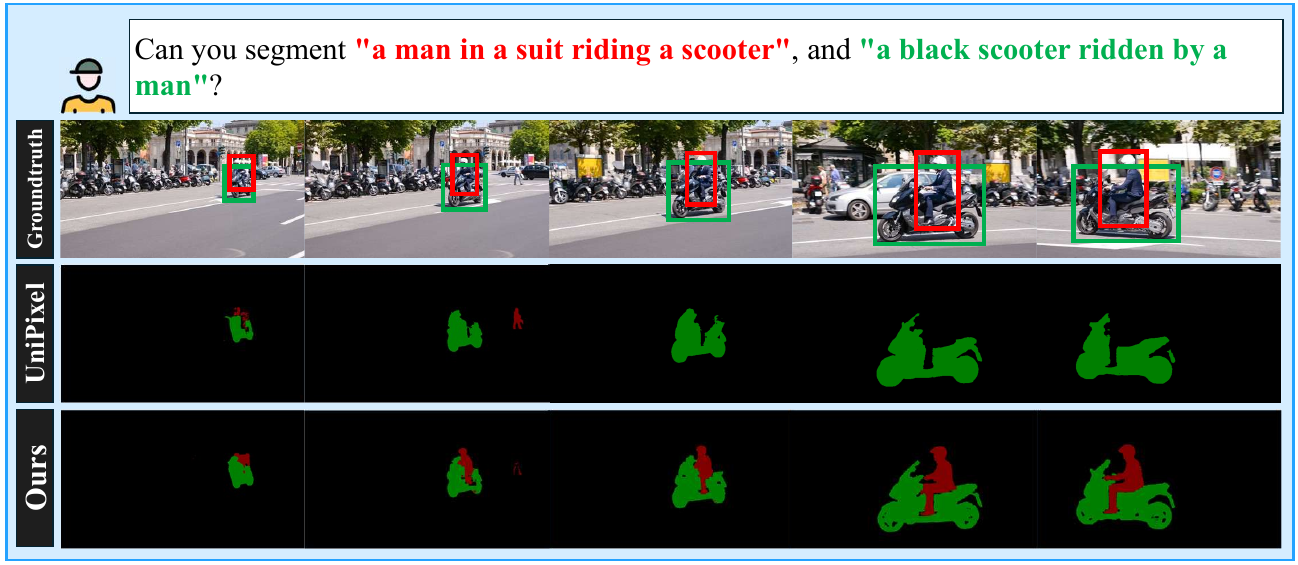}
\caption{
\textbf{Qualitative comparison on Ref-DAVIS17 (RVOS).}
For the expressions “a man in a suit riding a scooter” and “a black scooter ridden by a man,” our method accurately grounds both the rider and scooter from the first frame and maintains stable, instance-consistent masks throughout the sequence, preserving sharp boundaries despite significant motion and viewpoint changes. In contrast, UniPixel produces fragmented masks, inconsistent localization, and frequent structural loss. Our approach demonstrates stronger appearance-based grounding and greater robustness to multi-object consistency across frames.
}
\label{fig:davis_unipixel}
\end{figure*}

\noindent \textbf{Ref-DAVIS17.} Fig. \ref{fig:davis_videoglamm} shows a multi-query example involving several goldfish described by appearance and position (largest, smallest, center, end, bottom). VideoGLaMM produces overlapping masks and inconsistent instance separation, struggling to honor size- and location-based cues when the fish cluster tightly or share similar color. SPARROW cleanly distinguishes all five targets, maintaining boundaries even under occlusion and subtle pose changes. This demonstrates the model’s ability to resolve fine-grained appearance references without collapsing instances.

Fig. \ref{fig:davis_glus} presents another multi-object case with three expressions: “a back shooting gun,” “a black man,” and “a rope.” GLUS frequently produces masks that bleed across object boundaries, especially for thin structures such as the gun barrel and rope, and it occasionally merges the person with surrounding objects. SPARROW delivers sharp, stable masks for all three referents. The gun remains well-defined even during muzzle-flash frames, the person is consistently localized through large pose changes, and the rope retains its structure without fragmentation. This example highlights SPARROW’s improved precision on small or elongated objects.

Fig. \ref{fig:davis_unipixel} shows an example with two linked appearance-based queries: “a man in a suit riding a scooter” and “a black scooter ridden by a man.” UniPixel often produces fragmented masks, struggles to capture the full geometry of the scooter, and inconsistently localizes the rider during viewpoint changes. SPARROW maintains coherent masks for both the rider and the scooter across the entire clip, preserving structural details such as the rear wheel and handlebar region. The model consistently assigns each query to the correct instance despite motion, partial occlusion, and background variation, illustrating stronger appearance consistency and temporal stability.

\begin{figure*}[t]
\centering
\includegraphics[width=0.99\textwidth]{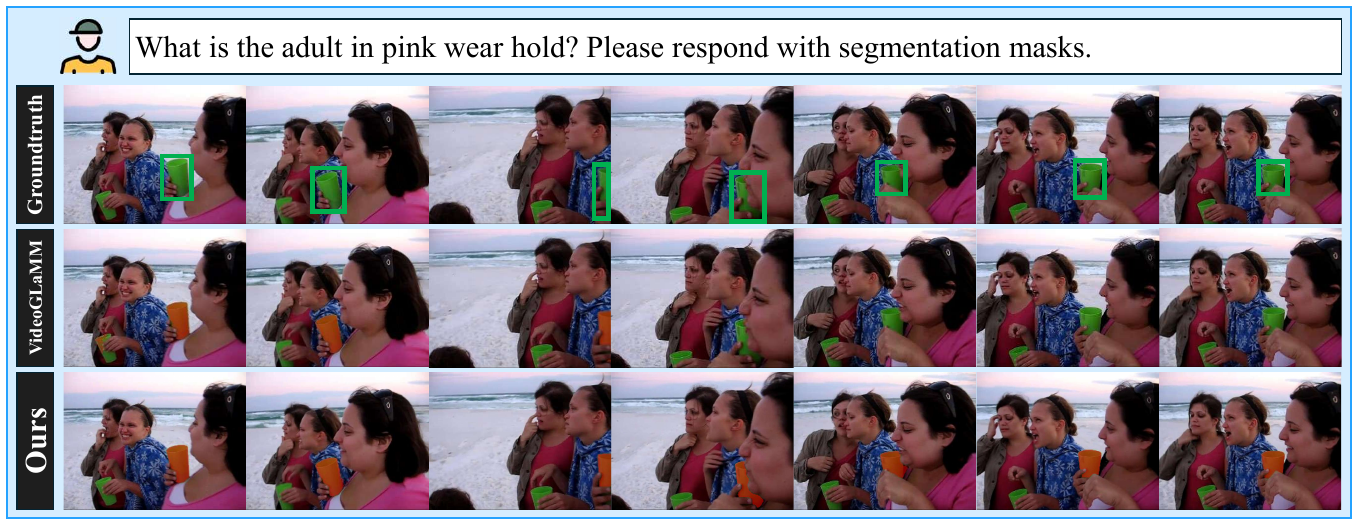}
\caption{
\textbf{Qualitative comparison on the VidSTG VG task.}
For the query “What is the adult in pink wear hold?,” our method consistently grounds the correct object—the orange cup—from the first frame and maintains accurate, stable masks despite hand motion, partial occlusions, and frequent interaction among multiple people. It preserves precise spatial extent and clean boundaries across the entire sequence.
In contrast, VideoGLaMM produces drifting or incomplete masks and struggles to maintain consistent localization of the held object during motion.
Our approach demonstrates strong spatio-temporal grounding and reliable object tracking on VidSTG.
}
\label{fig:vidstg_videoglamm}
\end{figure*}

\begin{figure*}[t]
\centering
\includegraphics[width=0.99\textwidth]{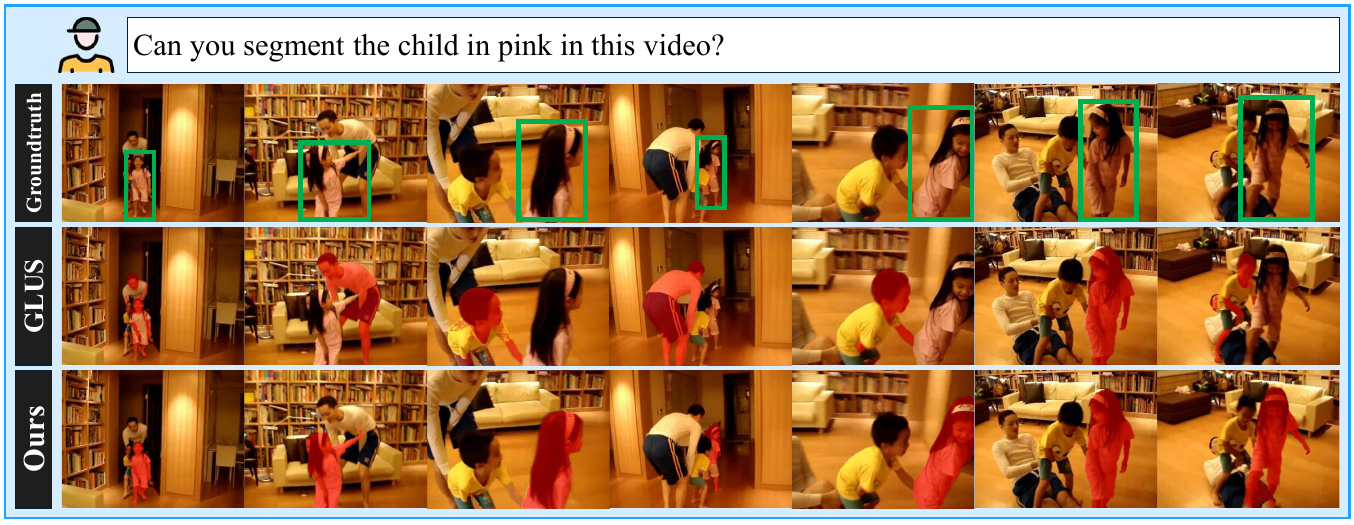}
\caption{
\textbf{Qualitative comparison on the VidSTG VG task.}
For the query “the child in pink,” our method accurately segments the correct child from the first frame and maintains stable masks throughout the sequence, preserving clean boundaries and consistent localization despite rapid motion, occlusions, and nearby interactions.
In contrast, GLUS often drifts to other people or yields incomplete, unstable masks under motion, whereas our approach remains robust to distractors and maintains stronger temporal consistency.
}
\label{fig:vidstg_glus}
\end{figure*}

\begin{figure*}[t]
\centering
\includegraphics[width=0.99\textwidth]{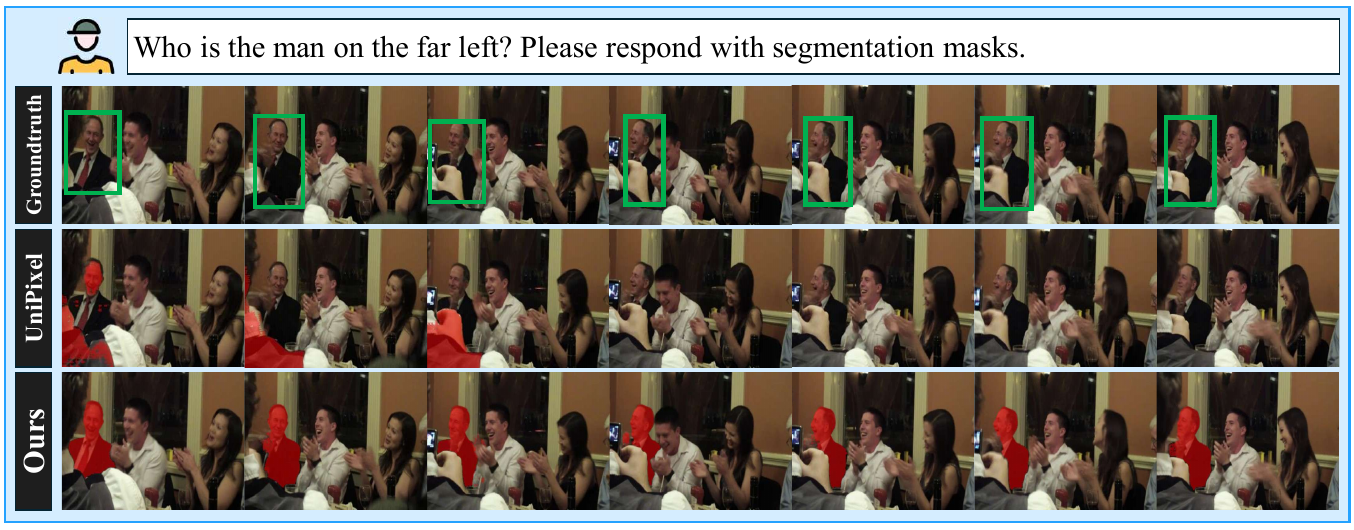}
\caption{
\textbf{Qualitative comparison on the VidSTG VG task.}
For the query “Who is the man on the far left?,” our method accurately grounds the correct individual from the first frame and preserves consistent identity assignments throughout the sequence, even during dense group interactions and frequent occlusions. It maintains stable masks and precise spatial localization across all frames.
In contrast, UniPixel exhibits identity drift and inconsistent mask assignment, especially in cluttered moments with overlapping people.
Our approach demonstrates stronger temporal grounding and robust identity preservation in challenging crowd scenarios.
}
\label{fig:vidstg_unipixel}
\end{figure*}

\subsection{Visual Grounding}
\noindent Fig. \ref{fig:vidstg_videoglamm} illustrates a grounding query from VidSTG: “What is the adult in pink wear hold?” The task requires identifying the adult in pink and localizing the object she is holding across the sequence. VideoGLaMM frequently produces incomplete or drifting masks, especially when the hands partially occlude the object or the viewpoint shifts. SPARROW consistently identifies the correct object, an orange cup, with clear boundaries and stable localization throughout the clip, demonstrating reliable spatio-temporal grounding under subtle hand–object interactions.

Fig. \ref{fig:vidstg_glus} presents a second grounding scenario involving the referent “the child in pink.” The scene includes multiple interacting children and adults, frequent occlusions, and rapid motion. GLUS exhibits noticeable drift, intermittently switching to nearby individuals or failing to capture the full extent of the target during pose changes. SPARROW maintains a stable, identity-consistent mask across the entire sequence, tracking the correct child even during occlusion and close-proximity interactions. This example highlights SPARROW’s robustness to distractors and improved temporal coherence in cluttered environments.

Finally, Fig.~\ref{fig:vidstg_unipixel} illustrates the query \textit{“Who is the man on the far left?”}, which requires maintaining identity under subtle pose changes in a crowded scene. UniPixel often shifts to adjacent individuals when poses or gestures are similar, causing identity switches and unstable boundaries. In contrast, SPARROW maintains the correct referent throughout, producing clean masks anchored to the intended individual despite crowd motion and visual ambiguity. This example highlights its robustness in identity-sensitive grounding with multiple similar candidates.

\begin{figure*}[t]
\centering
\includegraphics[width=0.99\textwidth]{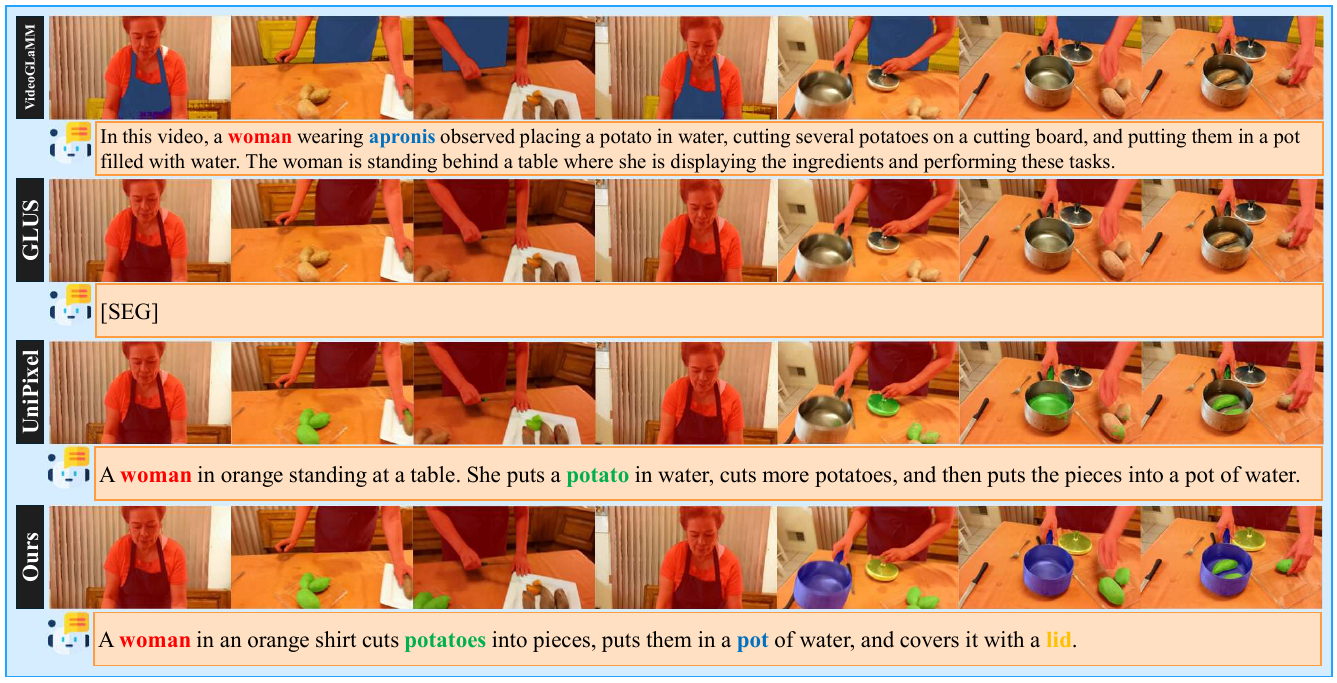}
\caption{
\textbf{Qualitative comparison on the Video GCG task.}
In a cooking sequence where a woman places a potato in water, chops additional potatoes, and transfers the pieces into a pot, our method maintains accurate pixel-level grounding of all manipulated objects across the full series of actions. It produces a coherent, temporally aligned description that faithfully reflects each step of the manipulation process.
In contrast, VideoGLaMM generates fluent but loosely grounded narratives, GLUS fails to produce a meaningful grounded response, and UniPixel captures only coarse actions while struggling with temporal transitions and hand–object interactions.
Our approach demonstrates robust spatio-temporal grounding and precise correspondence between language and visual evidence throughout the sequence.
}
\label{fig:gcg_1}
\end{figure*}

\subsection{Video Grounded Conversation Generation (Video GCG)}
\noindent Fig. \ref{fig:gcg_1} shows a clip where a woman in an orange shirt prepares ingredients by placing a potato in water, chopping more potatoes, and transferring them into a pot. VideoGLaMM generates a fluent narrative but frequently attributes actions that are not supported by the visual evidence, and its grounding does not reliably match the manipulated objects. GLUS fails to produce any grounded response, returning only \texttt{[SEG]}. UniPixel provides a coarse description but overlooks several action transitions and under-segments tool–hand interactions.
In contrast, SPARROW accurately follows the sequence, generating grounded text aligned with each sub-action and maintaining precise masks on hands, utensils, and ingredients. This example highlights SPARROW’s ability to couple fine-grained temporal actions with pixel-level grounding for grounded conversation generation.





\section{TSF Usage, Tradeoff, and Design}
\label{app:tsf}

\subsection{TSF Test-time Overhead}
\label{app:abs}

\noindent Relative to the baselines, enabling TSF adds (1) a single GroundingDINO pass on the first frame (batched multi-prompt inference), (2) a CLDTracker pass per frame per target, (3) CLIP ViT-L/14@336 crop-feature extraction per frame per target (batched on GPU), (4) lightweight K-means clustering per target ($K{=}4$ centroids), (5) a V$\!\rightarrow$L projection generating four TSF tokens per target, and (6) four additional LLM tokens per target. On an A100, the empirical latency is well approximated by \(\Delta t_{\text{TSF}}(n,T)\approx0.093+0.020\,n\,T\ \text{s}\), where $n$ and $T$ denote the number of tracked targets and frames, respectively. The detailed component-wise breakdown is summarized in Table~\ref{tab:tsf_overhead}. The cost scales linearly with both $n$ and $T$, dominated by CLDTracker and CLIP crops, while all other components are negligible ($<5$\,ms total). For $n{=}3$, $T{=}300$ (10\,s at 30\,FPS), the added time is $\approx18.1$\,s ($\sim$60\,ms/frame, or 20\,ms/target/frame).

\begin{table*}[t!]
\centering
\caption{\textbf{Component-wise analysis: TSF overhead at test time on A100.}}
\label{tab:tsf_overhead}
\setlength{\tabcolsep}{12pt}
\resizebox{0.82\textwidth}{!}{
\begin{tabular}{@{}lccc@{}}
\toprule
\textbf{Component} & \textbf{Per-unit latency} & \textbf{Count} & \textbf{Added time (s)} \\
\midrule
\multicolumn{4}{l}{\textbf{Vision + Tracking}} \\
GroundingDINO (first frame)      & 90 ms                     & $1$   & $0.090$ \\
CLDTracker                       & 18 ms / frame / target    & $nT$  & $0.018\,nT$ \\
CLIP ViT-L/14@336 (crops)        & 2 ms / crop (batched)     & $nT$  & $0.002\,nT$ \\
\midrule
\multicolumn{4}{l}{\textbf{Token construction + LLM}} \\
K-means clustering ($K{=}4$)     & $<1$ ms / target          & $n$   & $\approx 0$ \\
V$\!\to$L projection (4$n$ tok.) & 0.1 ms / token            & $4n$  & $0.0004\,n$ \\
LLM extra sequence (+4$n$ tok.)  & 0.5--1 ms / token         & $4n$  & $0.003\,n$ \\
\midrule
\textbf{Total} &  &  &
$\boxed{\Delta t_{\text{TSF}}(n,T)\approx0.093 + 0.020\,nT}$ \\
\bottomrule
\end{tabular}
}
\end{table*}

\subsection{Accuracy--Latency Tradeoff (Train-only vs.\ Train+Inference)}
\label{app:tsf_tradeoff}
\noindent The quantitative comparison of TSF usage modes is reported in the main paper. Training-only TSF improves $\mathcal{J\&F}$ by +2.9 without any inference-time tracking, indicating that pseudo-tracked supervision teaches target persistence and reduces identity switches; when combined with \texttt{[BOX]}, the gain increases to +7.3 while remaining overhead-free. Train+inference TSF achieves the highest score (77.7, +8.2) but incurs the additional tracking and token-construction overhead summarized in Table~\ref{tab:tsf_overhead}. The extra gain reflects the benefit of supplying temporally grounded identity cues directly during decoding, especially for longer sequences where geometric drift accumulates. In practice, Train-only TSF is the default choice for latency-sensitive settings, while inference-time TSF can be enabled when maximum temporal stability outweighs the overhead.

\begin{figure*}[!t]
\centering
\includegraphics[width=0.99\linewidth]{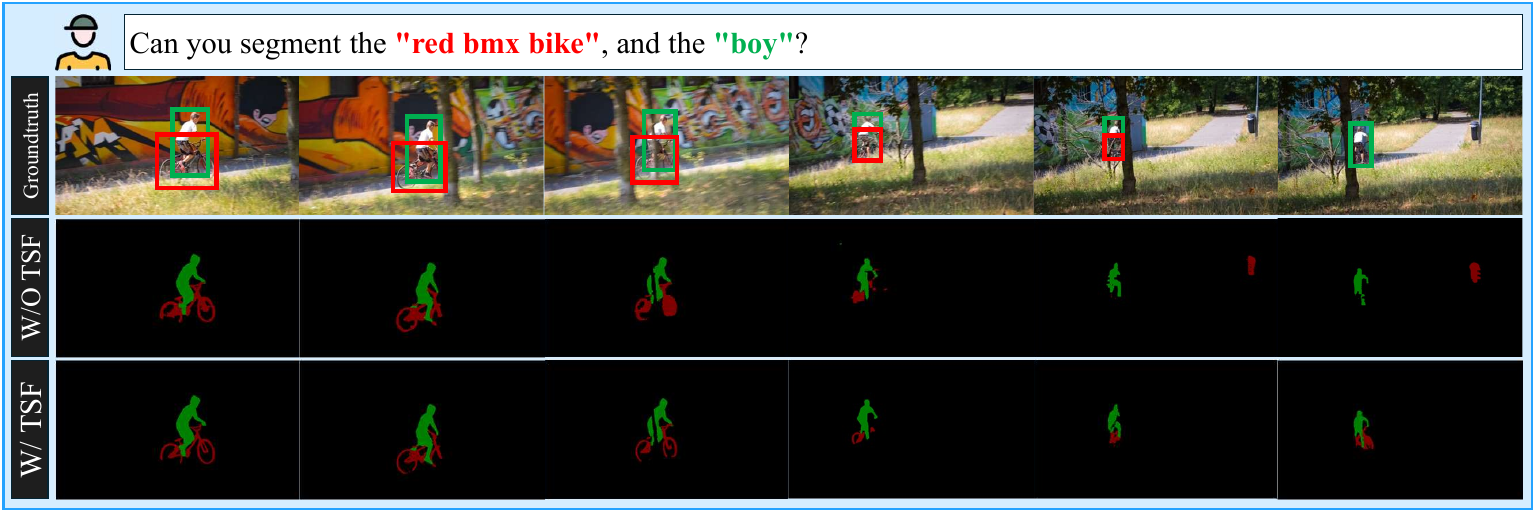}
\caption{\textbf{When to enable TSF at inference (qualitative).}
Query: segment the \textcolor{red}{``red bmx bike''} and the \textcolor{green}{``boy''}.
\emph{TSF Off (default)}: the small/fragile referent (bike) is prone to \emph{target dropout} and \emph{drift/false positives} as the target becomes small or partially occluded.
\textbf{TSF On (optional max-stability):} TSF provides a tracked, target-specific identity cue at inference, preserving consistent referent segmentation.}
\label{fig:tsf_infer_guidance}
\end{figure*}

\paragraph{When to Enable TSF at Inference}
\noindent By default we use \emph{Train-only TSF}, i.e., TSF is used during training but omitted at inference. TSF-at-inference is most beneficial under challenging temporal conditions: (i) occlusion and re-appearance, (ii) fast motion or motion blur, (iii) small or fragile targets, and (iv) crowded scenes with similar distractors. Fig.~\ref{fig:tsf_infer_guidance} shows a representative example where inference-time TSF reduces target dropout and drift. When enabled, the added cost is approximately $\sim$20\,ms per target per frame on A100 (Table~\ref{tab:tsf_overhead}). Practical guidance is summarized in Table~\ref{tab:tsf_when_enable}. This provides a simple accuracy--latency knob: keep TSF off at inference by default and enable it only when additional stability is critical.

\begin{table}[t]
\centering
\caption{\textbf{When to enable TSF at inference.}}
\label{tab:tsf_when_enable}
\setlength{\tabcolsep}{5pt}
\resizebox{0.99\columnwidth}{!}{
\begin{tabular}{ll}
\toprule
\textbf{Scenario} & \textbf{Recommendation} \\
\midrule
Short/simple clips, limited motion & Keep TSF off \\
Fast motion or motion blur & Enable TSF if accuracy is critical \\
Occlusion and re-appearance & Enable TSF \\
Small or fragile targets & Enable TSF if target dropout is likely \\
Crowded scenes with similar distractors & Enable TSF \\
\bottomrule
\end{tabular}
}
\end{table}

\subsection{TSF Injection Path Ablation}
\noindent In SPARROW, TSF tokens are extracted from tracked region crops, encoded by the spatial visual encoder, and projected into the LLM space through the spatial V--L projector. This design is intentional: TSF is intended to serve as an object-centric appearance and identity cue, and its feature distribution is therefore naturally aligned with the spatial stream. By contrast, the temporal V--L projector is optimized for global spatio-temporal video features rather than crop-level appearance features. To validate this design choice, we compare several TSF injection variants while keeping the rest of the pipeline unchanged: (i) \textbf{No TSF}, where no TSF tokens are injected; (ii) \textbf{spatial-only}, where TSF is projected only through the spatial V--L projector; (iii) \textbf{temporal-only}, where the same crop-based TSF features are projected through the temporal V--L projector; and (iv) \textbf{dual-path} variants, which combine both projectors either by concatenation or summation.

\begin{table}[!h]
\centering
\caption{\textbf{TSF injection path ablation on Ref-DAVIS17 (val).} We vary only the TSF projection path. ``tok.'' denotes the number of additional TSF tokens per target.}
\label{tab:tsf_proj_ablation}
\setlength{\tabcolsep}{6pt}
\resizebox{0.9\columnwidth}{!}{
\begin{tabular}{lcc}
\toprule
\textbf{TSF projection design} & \textbf{Extra tok.} & \textbf{J\&F $\uparrow$} \\
\midrule
No TSF & 0 & 72.5 \\
Temporal-only: $W_h(F_g(B_j))$ & +4 & 72.5 \\
Dual-path (concat): $[W_g(\cdot);\,W_h(\cdot)]$ & +8 & 74.9 \\
Dual-path (sum): $W_g(\cdot)+W_h(\cdot)$ & +4 & 74.7 \\
\rowcolor{blue!7.5} \textbf{Spatial-only (ours)}: $W_g(F_g(B_j))$ & +4 & \textbf{76.8} \\
\bottomrule
\end{tabular}
}
\end{table}

Table~\ref{tab:tsf_proj_ablation} shows that \textbf{spatial-only} performs best. Projecting crop-based TSF features through the temporal projector alone does not improve over the no-TSF baseline, suggesting that the temporal projection space is not well matched to object-centric supervision. Dual-path variants provide moderate gains, but remain below the spatial-only design, indicating that mixing projection spaces can introduce redundant or mismatched cues that weaken the identity signal.

\subsection{Robustness to Noisy TSF Supervision}
\label{app:robustness}
\noindent Since TSF supervision is constructed from pseudo-trajectories produced by GroundingDINO and CLDTracker, its quality may be affected by localization noise or occasional identity switches. We therefore stress-test robustness by corrupting only the training-time trajectories used for TSF extraction while keeping the model architecture, training schedules, and inference settings fixed.

\paragraph{Protocol.}
\noindent We evaluate on Ref-DAVIS17 (val) under the default Train-only TSF setting with \texttt{[BOX]} enabled. Only the teacher trajectories used to construct TSF tokens during training are corrupted; inference remains unchanged and does not use TSF tokens.

\paragraph{Corruption types.}
\noindent We consider three forms of corruption: (1) \textbf{Box jitter:} per frame, the box center is randomly shifted and its width/height are randomly rescaled by a factor controlled by $p\in\{5\%,10\%,20\%\}$. (2) \textbf{ID-switch injection:} per trajectory, we replace contiguous trajectory segments whose total length is $p\%$ of frames with another trajectory from the same video, simulating identity mismatches. (3) \textbf{Trajectory dropout:} TSF tokens are removed for a fraction $p\%$ of frames prior to TSF token construction.

\begin{table}[t]
\centering
\caption{\textbf{Robustness of SPARROW to noisy TSF supervision on Ref-DAVIS17 (val).}
We corrupt only the training-time teacher trajectories before TSF token extraction; the model architecture, training schedule, and inference setting remain unchanged.
All results use the default Train-only TSF setting with \texttt{[BOX]} enabled.}
\label{tab:tsf_stress}
\setlength{\tabcolsep}{6pt}
\resizebox{0.99\columnwidth}{!}{
\begin{tabular}{lcc}
\toprule
\textbf{Corruption applied to training trajectories} & \textbf{Level} & \textbf{J\&F $\uparrow$} \\
\midrule
None (clean TSF) & -- & 76.8 \\
\midrule
Box jitter (random shift/scale) & $\pm 5\%$ & 76.5 \\
& $\pm 10\%$ & 76.0 \\
& $\pm 20\%$ & 75.2 \\
\midrule
ID-switch injection (contiguous segments) & 5\% frames & 76.1 \\
& 10\% frames & 75.4 \\
& 20\% frames & 74.3 \\
\midrule
Trajectory dropout (remove TSF tokens) & 10\% & 76.6 \\
& 30\% & 76.0 \\
& 50\% & 75.0 \\
\midrule
No TSF baseline (\texttt{[BOX]} on) & -- & 72.5 \\
\bottomrule
\end{tabular}
}
\end{table}

\paragraph{Results.}
\noindent Table~\ref{tab:tsf_stress} shows that performance degrades gracefully as training-time TSF supervision becomes noisier. Under moderate corruption, SPARROW remains consistently above the \emph{No-TSF} baseline, indicating that TSF does not require perfectly accurate teacher trajectories to be beneficial. For example, with 10\% ID-switch injection, SPARROW achieves 75.4 J\&F, which remains +2.9 above the No-TSF baseline (72.5). These results suggest that the TSF mechanism is robust to realistic supervision noise.

\section{Comparison of Proposal Head Variants}
\label{app:proposal_head}
\noindent To justify our Transformer-based proposal module, we compare it with simpler alternatives under the same frozen-feature setting, where all methods operate on frozen SAM2 Hiera-L features. We evaluate: (i) \textbf{Direct LLM coordinates}, a proposer-free variant that predicts boxes directly from language outputs (Shikra-style); (ii) a \textbf{single-box MLP regressor} on pooled frozen visual features; and (iii) an \textbf{anchor-free dense convolutional head} (FCOS-style) with objectness, box offsets, and top-$K$ selection. For completeness, we also include \textbf{DETR} and \textbf{Deformable-DETR} decoder heads.

\begin{table}[h]
\centering
\caption{\textbf{Proposal generation variants (frozen SAM2 Hiera-L features).}
Performance on Ref-DAVIS17 (val) measured by $\mathcal{J\&F}$.
\emph{Latency is the incremental overhead of the proposal head only} (feature extraction shared).}
\label{tab:proposal_head_baselines}
\resizebox{0.99\columnwidth}{!}{
\begin{tabular}{lccc}
\toprule
Method & +Params (M)$\downarrow$ & +Latency (ms)$\downarrow$ & $\mathcal{J\&F}\uparrow$ \\
\midrule
Direct LLM coords (single box) & 0.00 & 0.0 & 67.8 \\
MLP regressor (single box) & 0.35 & 0.4 & 68.5 \\
Anchor-free conv head & 1.90 & 1.2 & 70.4 \\
DETR decoder head & 12.33 & 6.2 & 71.2 \\
\rowcolor{blue!7.5} Deformable-DETR decoder head (ours) & 11.38 & 5.6 & 72.5 \\
\bottomrule
\end{tabular}
}
\end{table}

\paragraph{Performance and efficiency.}
Table~\ref{tab:proposal_head_baselines} shows that direct regression and shallow heads underperform under frozen visual features, suggesting that this regime requires stronger proposal reasoning than a lightweight box predictor can provide. The anchor-free dense head improves over direct regression but remains below Transformer decoders in $\mathcal{J\&F}$. DETR-style proposers perform best overall, consistent with stronger class-agnostic objectness modeling and improved proposal coverage. Among all variants, \textbf{Deformable-DETR} achieves the best accuracy--efficiency tradeoff, improving over DETR while slightly reducing parameters and latency. Notably, the added parameters remain negligible compared to the MLLM's billion-param scale.

\paragraph{Proposal-set recall analysis.}
Because the language-conditioned filtering stage can only select from the available proposals, any ground-truth object missed at this stage is unrecoverable. We therefore evaluate oracle recall, defined as whether any of the top-$K$ predicted boxes overlap a ground-truth box above an IoU threshold.

\begin{table}[!h]
\centering
\caption{\textbf{Proposal-set oracle recall} under frozen SAM2 Hiera-L features.
Recall@IoU measures whether any of the top-$K$ boxes overlaps a GT box above the threshold.}
\label{tab:proposal_recall}
\resizebox{0.99\columnwidth}{!}{
\begin{tabular}{lccc}
\toprule
Method & $K$ & Recall@0.5$\uparrow$ & Recall@0.75$\uparrow$ \\
\midrule
Direct LLM coords (single box) & 1 & 50.0 & 20.0 \\
MLP regressor (single box) & 1 & 55.0 & 28.0 \\
Anchor-free conv head & 300 & 86.0 & 60.0 \\
DETR decoder head & 300 & 90.0 & 65.0 \\
\rowcolor{blue!7.5} Deformable-DETR decoder head (ours) & 300 & 93.0 & 70.0 \\
\bottomrule
\end{tabular}
}
\end{table}

\noindent As shown in Table~\ref{tab:proposal_recall}, Transformer-based proposers achieve higher oracle recall than regression or dense prediction heads, yielding a higher-coverage candidate set for language-conditioned selection and refinement. This matches our design goal: the proposer maximizes candidate coverage, while the language module focuses on selecting and refining the correct region rather than compensating for missed detections.

\section{Compute, Training, and Overhead}
\label{app:compute}

\subsection{Stage-wise Cost Analysis}
\noindent SPARROW introduces two additional components relative to baseline video-MLLM pipelines: (i) a one-time offline TSF supervision generation stage and (ii) a short two-stage fine-tuning procedure. Using the latency model in Appendix~\ref{app:abs}, processing a typical clip ($n{=}1$, $T{=}300$) for TSF generation requires approximately $\sim6.1$ s (Table~\ref{tab:tsf_overhead}). This preprocessing step is executed once, is not part of the training loop, and does not affect default inference.

Training proceeds in two stages: \textit{Stage 1 (TSF injection)}, where only multimodal adapters and LoRA parameters are optimized while the visual encoders and SAM2 decoder remain frozen; and \textit{Stage 2 (Dual-prompt box grounding)}, where the proposal generator is frozen and only the $[\texttt{BOX}]$ adapter and filtration head are updated. 

\begin{table}[!h]
\centering
\caption{\textbf{Training cost breakdown for SPARROW.} Offline TSF generation is one-time preprocessing and is not part of the training loop or default inference pipeline.}
\label{tab:training_cost}
\setlength{\tabcolsep}{5pt}
\resizebox{0.99\columnwidth}{!}{
\begin{tabular}{lccc}
\toprule
\textbf{Component} & \textbf{Wall-clock} & \textbf{GPU-hours} & \textbf{A100 GPU-days} \\
\midrule
Offline TSF precomputation (30,646 clips) & 51.9 h (1$\times$A100) & 51.9 & 2.16 \\
Baseline fine-tuning (per backbone) & 10 h (8$\times$A100) & 80 & 3.33 \\
Stage 1: TSF injection & 12 h (8$\times$A100) & 96 & 4.00 \\
Stage 2: Dual-prompt box learning & 2 h (8$\times$A100) & 16 & 0.67 \\
\midrule
\textbf{SPARROW training (S1 + S2)} & \textbf{14 h} & \textbf{112} & \textbf{4.67} \\
\textbf{Increment over baseline} & -- & \textbf{+32} & \textbf{+1.34} \\
\bottomrule
\end{tabular}
}
\end{table}

As shown in Table~\ref{tab:training_cost}, the combined SPARROW training (Stages 1+2) requires 112 GPU-hours (4.67 A100 GPU-days), corresponding to a modest +32 GPU-hours over baseline fine-tuning. To facilitate reproducibility, we release precomputed TSF trajectories and tokens, eliminating the need to rerun detection and tracking.

\begin{table}[!h]
\centering
\caption{\textbf{Execution summary across preprocessing, training, and inference.}}
\label{tab:cost_summary}
\setlength{\tabcolsep}{5pt}
\resizebox{0.99\columnwidth}{!}{
\begin{tabular}{lccc}
\toprule
\textbf{Component} & \textbf{Stage} & \textbf{When executed} & \textbf{Runtime cost} \\
\midrule
GroundingDINO + CLDTracker & TSF generation & Offline (one-time) & $\sim$6.1 s / clip \\
Crop encoding + K-means & TSF generation & Offline (one-time) & included in total TSF cost \\
\midrule
TSF Injection (Stage 1) & Training & Per backbone & $\sim$4.0 GPU-days \\
Dual-Prompt Learning (Stage 2) & Training & Per backbone & $\sim$0.67 GPU-days \\
\midrule
SPARROW inference (default) & Inference & Per video & no external detector/tracker \\
TSF-enabled inference (optional) & Inference & Per frame & $\sim$20 ms / target / frame \\
\bottomrule
\end{tabular}
}
\end{table}

Table~\ref{tab:cost_summary} clarifies when each component is executed. Heavy tracking modules (GroundingDINO and CLDTracker) are confined to the one-time TSF generation stage and are never invoked during default inference, preserving the deployment characteristics of the underlying backbone. TSF-at-inference remains an optional mode that adds $\sim$20\,ms per target per frame (Table~\ref{tab:tsf_overhead}).

\subsection{Inference Overhead}
\noindent We evaluate the inference overhead of SPARROW under identical evaluation settings for each backbone. Unless otherwise stated, all measurements are obtained on a single A100 GPU with batch size 1, using the same input resolution, frame sampling strategy, clip length, decoding settings, and evaluation pipeline as in the main experiments.

\paragraph{Inference setting.}
\noindent By default, SPARROW does not use TSF during inference. Accordingly, GroundingDINO, CLDTracker, CLIP-based crop processing, and K-means selection are not executed at test time. The resulting runtime overhead comes only from the added dual-prompt modules and the class-agnostic proposal head.

\paragraph{Reported metrics.}
\noindent We report three quantities: (i) end-to-end throughput in frames per second (FPS), (ii) vision GFLOPs per frame for the modules executed at inference, and (iii) total parameter count. GFLOPs/frame includes the visual backbone/encoder, mask decoder, and SPARROW heads/proposer, but excludes autoregressive LLM token generation, as its cost depends on the output length. End-to-end FPS captures the full runtime impact, including decoding, and is measured by averaging over the Ref-DAVIS17 (val) evaluation pipeline.

\begin{table}[!h]
\centering
\caption{\textbf{Inference overhead of SPARROW under identical evaluation settings.} We report total parameters (in Billion), end-to-end FPS, and vision GFLOPs per frame. +SP denotes the backbone augmented with SPARROW modules.}
\label{tab:inference_overhead}
\setlength{\tabcolsep}{5pt}
\resizebox{0.99\columnwidth}{!}{
\begin{tabular}{lrrrrrr}
\toprule
\textbf{Backbone} & \textbf{Params} & \textbf{FPS} & \textbf{GFLOPs/f} & \textbf{+SP Params} & \textbf{+SP FPS} & \textbf{+SP GFLOPs/f} \\
\midrule
VideoGLaMM & 5.435 & 2.40  & 38927 & 5.452 & 2.40  & 38962 \\
UniPixel   & 3.881 & 15.38 & 1530  & 3.898 & 15.04 & 1565 \\
GLUS       & 7.288 & 6.374 & 2680  & 7.305 & 6.29  & 2715 \\
\bottomrule
\end{tabular}
}
\end{table}

\paragraph{Results.}
\noindent Table~\ref{tab:inference_overhead} shows that SPARROW introduces only modest inference overhead across all three backbones. The parameter increase is limited to +0.017B, while the throughput reduction is minor: VideoGLaMM is unchanged within measurement noise (2.40$\rightarrow$2.40 FPS), UniPixel drops from 15.38 to 15.04 FPS, and GLUS drops from 6.374 to 6.29 FPS. The increase in GFLOPs/frame is similarly small, indicating that the performance gains are not driven by substantial additional test-time computation.

\appendix

\end{document}